\documentclass[letterpaper]{article} 
\usepackage{aaai2027}  
\usepackage[hyphens]{url}  
\usepackage{graphicx} 
\usepackage{natbib}  
\usepackage{caption} 
\usepackage{algorithm}
\usepackage{algorithmic}
\usepackage{appendix}
\usepackage{booktabs}
\usepackage{multirow}
\usepackage{graphicx}
\usepackage{enumitem}
\usepackage{amsthm,amsmath,amssymb}
\usepackage{mathrsfs}

\newcommand{\subtitle}[1]{{\noindent}{\textbf{#1}}}

\makeatletter
\g@addto@macro\@maketitle{%
  \par\medskip
  \centering
  \includegraphics[width=\textwidth]{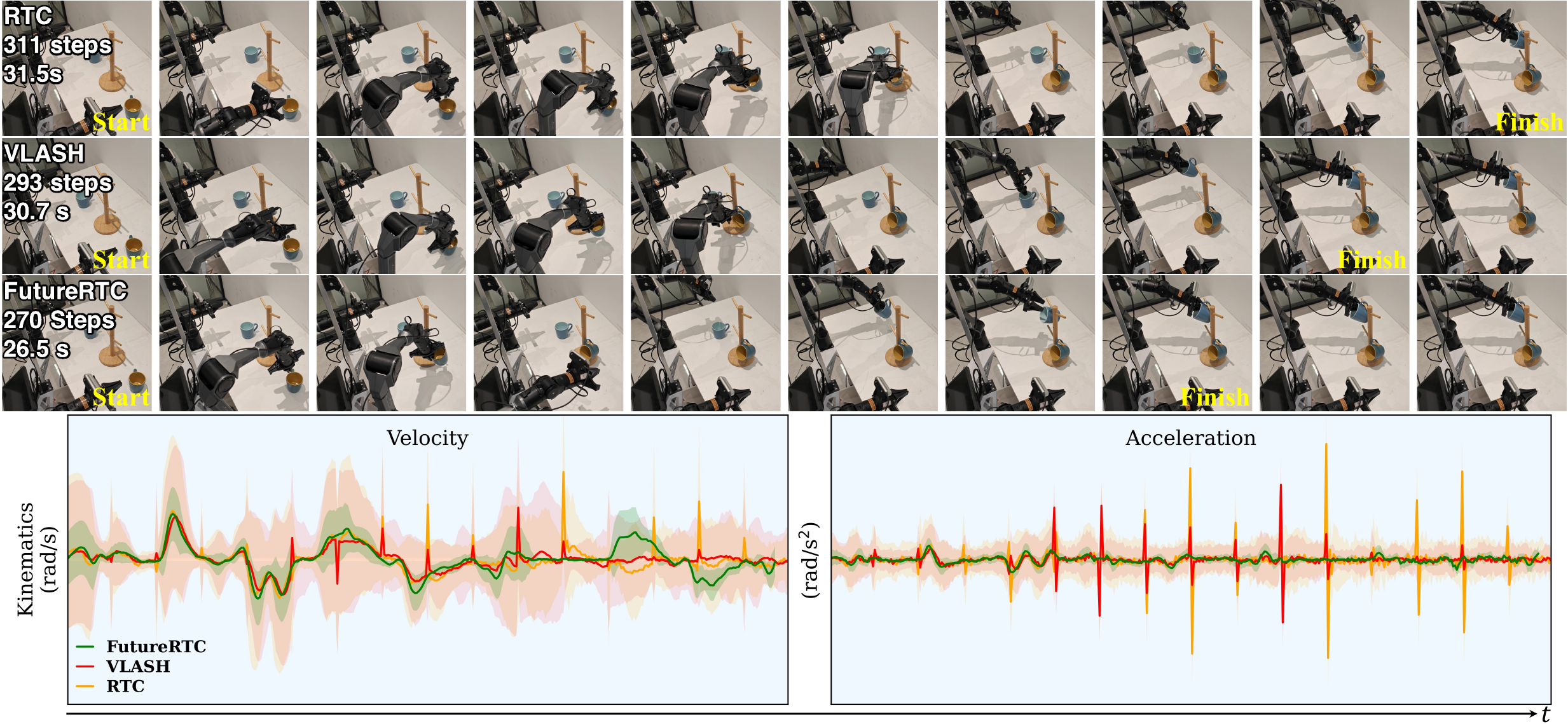}%
  \captionof{figure}{Comparison of the inference-time method RTC~\cite{RTC}, training-time method VLASH~\cite{VLASH}, and our FutureRTC on the real-world ``\emph{Hang Cups}'' task with $\pi_{0.5}$~\cite{pi05} as the VLA backbone. \textbf{Top}: Task completion progress under asynchronous execution. \textbf{Bottom}: Average robot kinematics during task execution.}
  \label{fig: teaser}
  \medskip
}
\makeatother

\usepackage{newfloat}
\usepackage{listings}
\DeclareCaptionStyle{ruled}{labelfont=normalfont,labelsep=colon,strut=off} 
\floatstyle{ruled}
\newfloat{listing}{tb}{lst}{}
\floatname{listing}{Listing}

\usepackage{booktabs}
\usepackage[pagebackref,breaklinks,colorlinks,allcolors=blue]{hyperref}
\title{FutureRTC: Real-Time Robot Execution with \\ Anticipatory-Conditioned Action Chunking}
\author{
    Hai Jiang\textsuperscript{\rm 1},
    Yixian Zou\textsuperscript{\rm 2},
    Binbin Liang\textsuperscript{\rm 1},
    Boqian Liu\textsuperscript{\rm 3},
    Fanman Meng\textsuperscript{\rm 2},
    Shuaicheng Liu\textsuperscript{\rm 2}\corresponding
}
 
\affiliations{
    \textsuperscript{\rm 1}School of Aeronautics and Astronautics, Sichuan University\\
    \textsuperscript{\rm 2}University of Electronic Science and Technology of China \\
    \textsuperscript{\rm 3}University of Alberta \\
    jianghai@stu.scu.edu.cn, \{zouyixian@std., liushuaicheng@\}uestc.edu.cn
}

\begin{document}

\maketitle

\begin{abstract}
Real-time deployment of Vision-Language-Action (VLA) policies necessitates asynchronous execution, wherein subsequent action chunks are computed concurrently with the execution of the current chunk, leading to prediction-execution misalignment and manifesting as inter-chunk discontinuities. Existing methods either superficially smooth chunk boundaries, require costly policy optimization, or exclusively forward-predict proprioceptive states yet neglect critical visual observations. In this paper, we propose \textbf{FutureRTC}, a plug-and-play adaptation framework that predicts execution-time observations and states for asynchronous VLA control without modifying the underlying policy. Specifically, FutureRTC features a state correction module to compensate for the discrepancy between rolled-forward and actual execution-time proprioceptive states and an observation prediction module that forecasts execution-time visual representations by leveraging robot motion as an explicit physical prior through motion-aware feature transport and reconstruction. Furthermore, we introduce a policy consistency loss to align the action chunks generated from predicted contexts with those produced under the expected execution-time inputs of the VLA policy. Extensive experiments across simulated and real-world environments demonstrate that FutureRTC achieves superior robustness to inference delays, resulting in smoother trajectories, faster execution, and consistently higher task success rates. \href{https://jianghaiscu.github.io/FutureRTC_proj/}{Project Website}.
\end{abstract}

\section{Introduction}\label{sec: introduction}
Vision-Language-Action (VLA) models have recently emerged as an attractive paradigm for robot learning by unifying deep visual-linguistic reasoning with expressive action generation~\cite{OpenVLA, pi0, pi05}. To reconcile the substantial computational cost of VLA inference with the high-frequency requirements of low-level robot control, \emph{action chunking}~\cite{ACT} has become a widely adopted strategy, which enables the policy to generate a sequence of future actions in a single forward pass, thereby amortizing the inference overhead through open-loop execution. However, as VLA models continue to scale in size and are deployed on resource-constrained edge platforms or remote servers, inference latency has become a fundamental challenge for real-time deployment~\cite{RTC, real_time_vla_v1}. In standard \emph{synchronous execution}, the robot is forced to idle during chunk computation, resulting in stop-and-go behaviors that severely degrade real-time responsiveness and manipulation precision.

To circumvent execution stalls, recent systems adopt \textit{asynchronous execution}~\cite{RTC}, wherein the policy generates future action chunks in parallel with the execution of the current chunk. While improving computational efficiency, conducting mutations at chunk boundaries causes \emph{inter-chunk discontinuities} that accumulate errors and lead to task failures. Existing efforts to mitigate discontinuities generally fall into inference-time and training-time strategies. The former~\cite{ACT, BID, RTC, PAINT} utilize heuristic post-processing to smooth chunk boundaries without retraining the policy, often introducing additional computational overhead. In contrast, most training-time methods~\cite{training_rtc, REMAC, DiscreteRTC} explicitly adapt the policy to asynchronous execution, while requiring costly retraining or fine-tuning the VLA model. 

However, neither of the above approaches resolves the fundamental \emph{prediction-execution misalignment}, where the executed actions remain conditioned on observations and proprioceptive states from an earlier time step, causing them to deviate from the evolving scene as inference delay grows. VLASH~\cite{VLASH} attempts to reduce the misalignment by forward-predicting the execution-time proprioceptive state while still relying on outdated visual observations. Since the visual scene evolves alongside the robot configuration during the delay interval, correcting the state alone is insufficient to provide the execution time context required by the policy. As shown in Fig.~\ref{fig: teaser}, both the inference-time method RTC and the training-time method VLASH exhibit jerky trajectories and slower task completion speed.

To this end, we propose \textbf{FutureRTC}, a plug-and-play adaptation framework that anticipates the execution-time observation and state that the policy should condition on to address prediction-execution misalignment for real-time robotic control while leaving the base VLA untouched. Specifically, we first introduce a state correction module that compensates for the residual discrepancy between the state obtained by rolling the committed actions forward and the actual execution-time proprioceptive state. Subsequently, we observe that correcting the state alone is insufficient, as the stale observation is itself a critical factor whose neglect causes substantial performance degradation that intensifies as the inference delay increases. To this end, we propose an observation prediction module that predicts the execution-time visual representation from the stale observation, committed actions, and corrected state. By leveraging robot motion as the physical prior, the module spatially transports existing visual features through motion-aware warping and reconstructs newly exposed content during the delay interval. Furthermore, we introduce a policy consistency loss to encourage the action chunk generated by the untouched policy from the predicted observation and corrected state to match the one generated from the actual execution-time pair. As shown in Fig.~\ref{fig: teaser}, our method presents smoother trajectories and faster task completion speed. Extensive experiments across simulated and real-world settings demonstrate that our FutureRTC delivers faster execution, robust delay tolerance, and consistently higher success rates.

Our contributions can be summarized as follows:
\begin{itemize}
    \item We propose FutureRTC, a plug-and-play adaptation framework that generates execution-time observations and states to mitigate asynchronous prediction-execution misalignment for real-time robot execution.
    \item We propose a state correction module to compensate for the residual discrepancy between rolled-forward and actual execution-time states, and an observation prediction module to forecast execution-time visual observations, using a policy consistency loss to align the predicted context with the distribution expected by the untouched policy.
    \item Extensive evaluations demonstrate that FutureRTC achieves superior robustness under varying inference delays, smoother trajectories, faster task completion, and consistently improved task success rates.
\end{itemize}

\begin{figure*}[!ht]
    \centering
    \includegraphics[width=\linewidth]{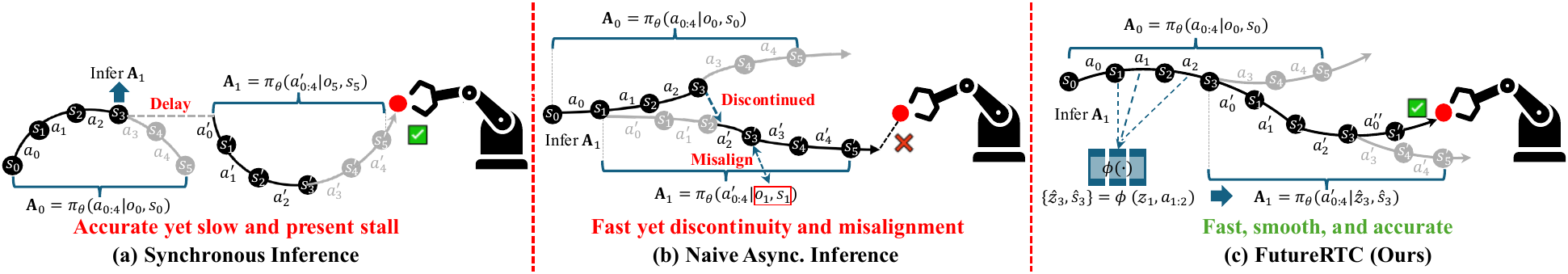}
    \caption{Comparison between our method and existing execution paradigms, illustrated
  with prediction horizon $H=5$, execution horizon $K=3$, and inference delay $d=2$. \textbf{(a) Synchronous:} The robot stalls during inference, yielding accurate but stop-and-go motion. \textbf{(b) Naive Asynchronous:} Chunk $\mathbf{A}_1$ is conditioned on the stale observation-state pair $(o_1,s_1)$ yet executed at step $3$, suffering from prediction-execution misalignment and inter-chunk discontinuities. \textbf{(c) FutureRTC (Ours):} Our plug-and-play adapter $\phi(\cdot)$ forecasts the execution-time context $(\hat z_3,\hat s_3)=\phi(z_1,a_{1:2})$ from the stale $o_1$ and committed actions $a_{1:2}$, where $z$ is the latent feature of the observation $o$ obtained by the vision encoder of the pretrained VLA. Conditioning the policy on this anticipated pair promotes continuous, reactive, and smooth execution.}
    \label{fig: inference_compare}
\end{figure*}

\section{Related Work}\label{sec: related_work}
\subsection{VLA and Action Chunking}\label{subsec: vlas}
Vision-Language-Action models have advanced robot learning by integrating visual perception, language understanding, and action prediction within a unified policy framework~\cite{vla_survey}. Representative methods~\cite{OpenVLA, pi0, Rdt-1b, Gr00t, Smolvla, pi05, Dexvla, Acot-vla, LingBotVLA} leverage large-scale multimodal pre-training and integrate deep visual-linguistic representations with expressive action generation paradigms, including diffusion models and flow matching~\cite{flow_matching, DP3, DP}, to improve cross-task generalization. However, the sheer scale of these models, with substantial computational costs, makes inference latency a critical bottleneck for real-time robotic control. In synchronous control loops, such latency forces robots to wait for the next action prediction, thereby reducing closed-loop responsiveness in dynamic scenes.

To alleviate the computational bottleneck and mitigate the compounding errors associated with step-wise execution, action chunking has been widely adopted as the standard paradigm in VLAs. By generating multi-step action sequences in parallel, it bridges the gap between low-frequency policy inference and high-frequency robot control. However, open-loop execution of long chunks weakens short-term responsiveness to environmental perturbations, whereas executing shorter chunks often introduces boundary discontinuities and jerky motions. Therefore, balancing temporal consistency with environmental responsiveness remains a central challenge in chunk-based VLA deployment.

\subsection{Asynchronous VLA Inference}\label{subsec: action_chunk}
To mitigate execution stalls, recent systems adopt asynchronous inference, in which the robot continues executing the remaining actions of the previously generated chunk while the policy computes the next chunk in parallel, which may amplify inter-chunk discontinuities and introduce synchronization challenges between prediction and execution. Existing remedies fall into inference-time and training-time strategies. The former~\cite{ACT, RTC, BID, PAINT} operate only during inference by modifying the generated actions or execution procedure without updating the policy parameters, but often introduce additional runtime computation, thus making it undesirable for low-latency robotic control. In contrast, training-time methods~\cite{training_rtc, A2C2, REMAC, Legato, DiscreteRTC} require extra optimization stages, by fine-tuning, retraining, or augmenting the VLA policy with trainable modules, to explicitly adapt the policy to asynchronous execution. Nevertheless, asynchronous inference inherently introduces prediction-execution misalignment, as actions are generated from observations that are stale before execution. VLASH~\cite{VLASH} alleviates this issue by conditioning on future proprioceptive states, but it still relies on outdated visual observations and therefore remains insensitive to environmental changes. In this paper, we present an asynchronous inference framework in which a plug-and-play adapter is constructed to predict execution-time observations and states from their stale counterparts, thereby enabling smoother, faster, and more accurate execution.

\section{Preliminaries}\label{sec: preliminaries}
\subtitle{Action Chunking Policy.} For a given control timestep $t$, the VLA policy maps the visual observation $o_t$ and proprioceptive state $s_t$ to a trajectory of $H$ future actions as:
\begin{equation}\label{eq:1}
\mathbf{A}_m=(a_t,a_{t+1},\dots,a_{t+H-1})\sim\pi_\theta(\cdot\mid o_t,s_t),
\end{equation}
where $m$ indexes successive chunks and $H$ represents the \emph{prediction horizon}~\cite{ACT}. To circumvent the computational burden of per-step replanning, the controller executes a subset of $K$ actions ($K \le H$) from $\mathbf{A}_m$ prior to the next policy query, where $K$ denotes the \emph{execution horizon}. While such a chunking paradigm mitigates inference costs and promotes temporal consistency, it introduces a structural dependency between the control loop frequency and sequence generation latency.

\subtitle{Flow-matching Generation.} In line with recent VLA models~\cite{pi0, pi05, Smolvla}, we parameterize the action generation process of our policy via flow matching~\cite{flow_matching}. To generate an action chunk, an initial latent state $\mathbf{A}_m^{0}\sim\mathcal{N}(\mathbf{0},\mathbf{I})$ is transformed by integrating a learned velocity field $v_\theta$ over a flow-matching time $\tau\in[0,1]$ using $n$ Euler steps as:
\begin{equation}
\mathbf{A}_m^{\tau+\frac{1}{n}}=\mathbf{A}_m^{\tau}+\frac{1}{n}\,v_\theta(\mathbf{A}_m^{\tau},o_t,s_t,\tau),
\end{equation}
yielding the final iterate $\mathbf{A}_m^{1}$ as the predicted action sequence.

\subtitle{Inference Delay.} Generating an action chunk incurs a non-negligible wall-clock time $\delta$. Given a base controller period $\Delta t$, we define the \emph{inference delay} as $d=\left\lfloor \delta/\Delta t \right\rfloor$, which quantifies the number of control steps that elapse between acquiring the observation-state pair $(o_t,s_t)$ and the chunk $\mathbf{A}_m$ being ready for execution~\cite{RTC}. As VLAs scale in model capacity and shift toward remote or edge-constrained inference, the latency $\delta$, and consequently the inference delay $d$, inevitably increases.

\subtitle{Synchronous and Asynchronous Execution.} In \emph{synchronous} execution, the control loop blocks during the generation of $\mathbf{A}_m$, while ensuring each action chunk is conditioned on the recent state $(o_t,s_t)$, it introduces robot stalls for $d$ steps between chunks, thereby breaking real-time control continuity, as shown in Fig.~\ref{fig: inference_compare} (a). \emph{Asynchronous} execution removes this stall by generating the subsequent chunk in the background. To ensure that the next chunk $\mathbf{A}_{m+1}$ is ready precisely when the current $K$ actions conclude, its inference must be initiated $d$ steps in advance. Consequently, the new chunk is conditioned on the stale $(o_{t+K-d},s_{t+K-d})$, yet executes at step $t+K$. The first $d$ actions in $\mathbf{A}_{m+1}$ would be discarded since they overlap with the committed tail of the current chunk, only the remainder $\mathbf{A}_{m+1}[d{:}d{+}K-1]$ is spliced on and executed, as shown in Fig.~\ref{fig: inference_compare} (b).

\subtitle{Prediction-Execution Misalignment.} Formulating the subsequent chunk as $\mathbf{A}_{m+1}=\pi_\theta(\cdot\mid o_{t+K-d},s_{t+K-d})$, the actions executed from step $t+K$ relies on an observation and state that are $d$ steps obsolete. This temporal gap between the conditioning time $t+K-d$ and the execution time $t+K$ constitutes a critical \emph{prediction-execution misalignment} that manifests as inter-chunk discontinuities. Furthermore, since the scene and robot would evolve during the $d$-step window, it causes the executed actions to deviate from the intended behavior, exacerbating the error as $d$ grows.

\section{Methodology}\label{sec: methodology} 
Given a flow-matching VLA $\pi_\theta(\cdot\mid o_t,s_t)$, our objective is to mitigate the discrepancy of asynchronous execution, wherein the policy is conditioned on the stale pair $(o_{t+K-d},s_{t+K-d})$ prior to its execution at step $t+K$. To rectify this prediction-execution misalignment and realize reliable closed-loop performance, we introduce \textbf{FutureRTC}, which incorporates a plug-and-play anticipatory adapter $\phi(\cdot)$ to forecast the execution-time observation and state from their stale counterparts, promoting highly reactive and kinematically smooth robotic control, as shown in Fig.~\ref{fig: inference_compare} (c).

\subsection{State Correction Module}
The execution-time state $s_{t+K}$ remains unobservable at the generation step $t+K-d$, while it can be estimated by rolling forward the stale state $s_{t+K-d}$ using the committed actions $[a_{t+K-d}: a_{t+K-1}]$ following~\cite{VLASH} as:
\begin{equation}
    \tilde{s}_{t+K} = s_{t+K-d} \oplus \sum_{i=t+K-d}^{t+K-1} a_i.
\end{equation} 
However, whether under absolute or relative posture control, forward integration inevitably accumulates bias due to the inherent deviation between the commanded motion and the actual robot motion. To this end, we introduce a State Correction Module (SCM) $\phi_\text{SCM}(\cdot)$ that consists of MLP layers, as shown in Fig.~\ref{fig: pipeline} (a), to predict a compensatory residual as $\tilde{s}_{\Delta} = \phi_\text{SCM}(\tilde{s}_{t+K},\,\frac{d}{d_{\max}})$, to complement the forward integration process. The final execution-time state is obtained as $\hat{s}_{t+K}=\tilde{s}_{t+K}\oplus\tilde{s}_{\Delta}$, with the orientation updated through relative-rotation composition, which serves simultaneously as the conditioning signal for the observation predictor and the state input for the VLA policy. During the training phase, we employ a state correction loss $\mathcal{L}_{\text{state}}$ to minimize the discrepancy between the predicted $\tilde{s}_{\Delta}$ and its ground-truth counterpart recorded in the demonstrations as:
\begin{equation}
\mathcal{L}_{\text{state}}=||(s_{t+K}\ominus\tilde{s}_{t+K}) - \tilde{s}_{\Delta}||_2^2.
\end{equation}

\subsection{Observation Prediction Module}
\begin{figure}[!t]
    \centering
    \includegraphics[width=\linewidth]{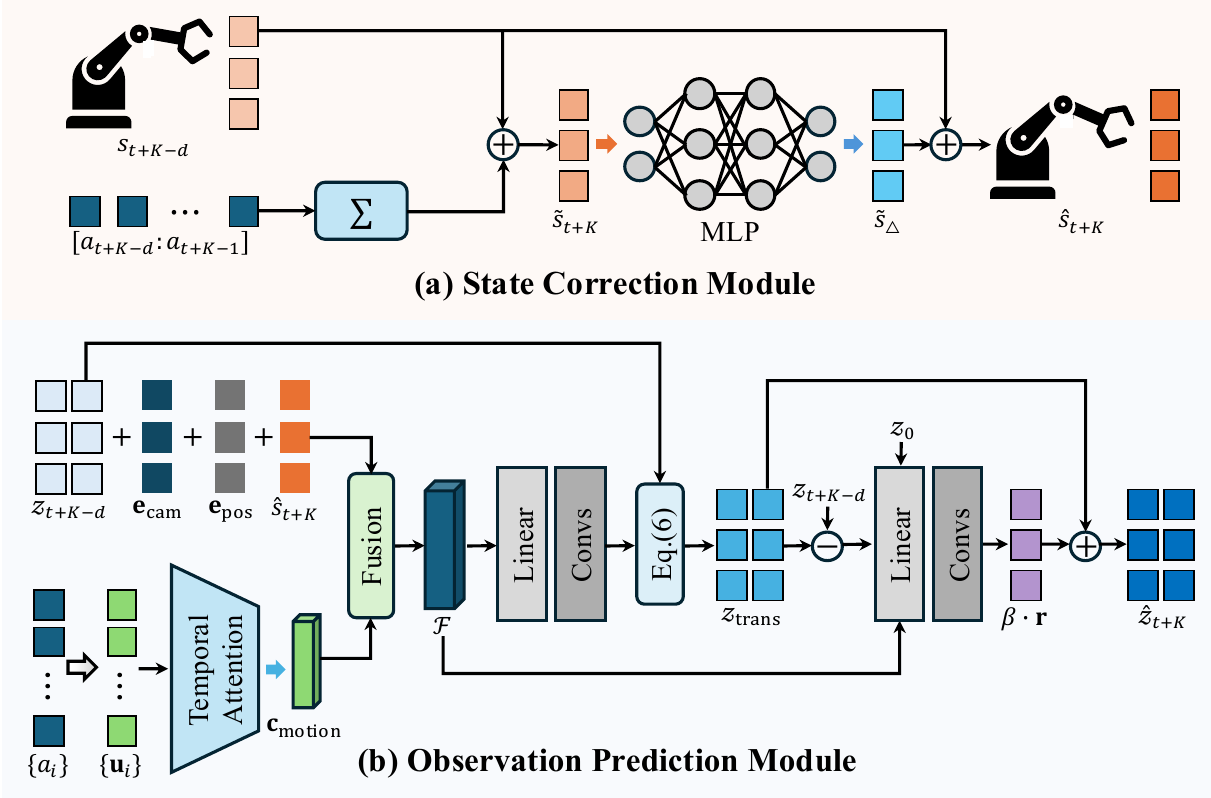}
    \caption{Illustration of the detailed architecture of the proposed adapter, which consists of a State Correction Module and an Observation Prediction Module to generate the execution-time observation and state from stale information.}
    \label{fig: pipeline}
\end{figure}

While the state correction module mitigates a fraction of the prediction-execution misalignment, observational staleness remains a critical unaddressed factor causing performance degradation, the severity of which increases sharply as the delay $d$ increases. To estimate the unobservable execution-time observation $o_{t+K}$, we introduce an Observation Prediction Module (OPM) $\phi_{\text{OPM}}(\cdot)$, as shown in Fig.~\ref{fig: pipeline} (b), which leverages the stale observations, the committed actions, and the corrected state to generate the execution-time observation. Instead of reconstructing high-dimensional pixels, our OPM operates within the latent space, i.e., the feature $z=\mathcal{E}(o)$ obtained from the vision encoder $\mathcal{E}(\cdot)$ of the pretrained VLA, ensuring the computational overhead is relatively limited compared to policy inference while generating representations that are inherently compatible with VLA policies.

Since the primary visual variation is induced by the physical displacement of the robot, we model feature evolution as spatial transportation rather than synthesizing it from scratch. The committed action sequence within the delay window $[a_{t+K-d}: a_{t+K-1}]$ dictates the physical motion responsible for scene changes and thus serves as an explicit motion prior indicating where existing content should be transported. However, feeding the action sequence directly would force the network to implicitly learn forward integration. We instead map the action trajectory into an explicit physical feature space. For each committed action $a_i$ within the window, we construct a physical feature vector $\mathbf{u}_i$ as:
\begin{equation}
    \mathbf{u}_i=\Big[\,a_i,\ \textstyle\sum_{j\le i} a_j,\ a_i-a_{i-1},\ \sum_{j\le i}|a_j|,\ \tfrac{i}{d}\,\Big],
\end{equation}
which respectively encode the instantaneous command, cumulative displacement, action increment, accumulated path length, and temporal progress. Concatenating the features across all steps yields a trajectory feature matrix $\mathbf{U}=[\mathbf{u}_1,\mathbf{u}_2,\dots,\mathbf{u}_d]^\top$, which is fed into a temporal self-attention block to capture temporal dependencies and produce the action-conditioned motion prior $\mathbf{c}_{\text{motion}}$.

\begin{figure}[!t]
    \centering
    \includegraphics[width=\linewidth]{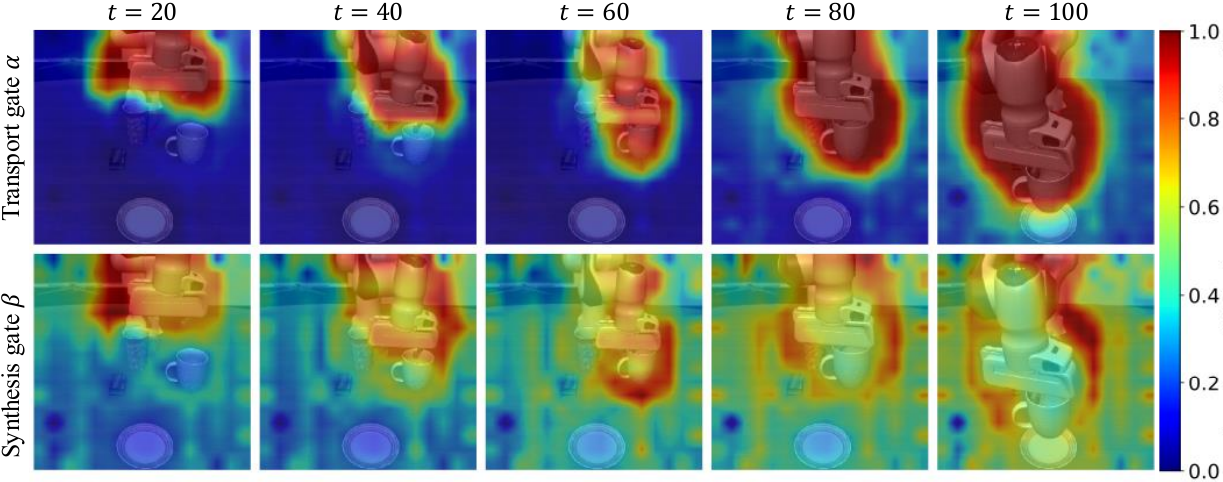}
    \caption{Illustration of the transport gate $\alpha$ (top) and synthesis gate $\beta$ (bottom) during execution. The transport gate activates on the moving arm and gripper, propagating existing visual features according to the committed motion, whereas the synthesis gate highlights motion boundaries and newly revealed regions that require feature synthesis.}
    \label{fig: gate_visualize}
\end{figure}

We then fuse the motion prior $\mathbf{c}_{\text{motion}}$, the stale latent $\mathbf{z}_{t+K-d}$ together with the per-camera embedding $\mathbf{e}_{\text{cam}}$ which distinguishes the camera streams and position embedding $\mathbf{e}_{\text{pos}}$, and the corrected state $\hat{s}_{t+K}$ to supply the actual end-effector configuration at execution time $t+K$ to form a conditioning feature $\mathcal{F}$. However, directly regressing the future latent from $\mathcal{F}$ is highly ill-posed and physically unconstrained. We therefore explicitly project $\mathcal{F}$ by cascaded linear and convolutional layers into a 2D flow displacement $\Delta\mathbf{p}$ and a transport gate $\alpha\in[0,1]$. The stale feature is spatially warped using differentiable bilinear grid sampling to pull in the motion-transported content and blend it back through a gated combination to obtain $z_{\text{trans}}$, whose content has been repositioned to the location carried by the motion flow, as:
\begin{equation}
z_{\text{trans}}=(1-\alpha) \cdot z_{t+K-d}+\alpha \cdot \mathcal{W}(z_{t+K-d},\ \mathbf{p}-\Delta\mathbf{p}),
\end{equation}
where $\mathcal{W}(\cdot)$ and $\mathbf{p}$ denote the bilinear grid sampling and the 2D coordinate grid, respectively. The transport gate $\alpha$ serves as a per-token weight that determines the contribution of warped features, where static regions preserve their original representations ($\alpha\rightarrow0$), while regions undergoing robot-induced motion increasingly rely on the warped features ($\alpha\rightarrow1$), as shown in the top row of Fig.~\ref{fig: gate_visualize}.

While warp transport handles coordinate-space displacement, it is fundamentally restricted to feature rearrangement and cannot synthesize unknown content, such as visual deformations upon gripper contact or newly revealed disoccluded regions, which cannot be explained by geometric warping. To this end, we introduce a gated residual branch consisting of several linear and convolutional layers to compensate for content unexplained by transport, which ingests the conditioning feature $\mathcal{F}$, the transport residual $z_{\text{trans}} -z_{t+K-d}$ that represents changes already resolved by advection, and the initial observation $z_{0}$ as a static visual reference from which disoccluded content can be borrowed, to predict a synthesis feature $\mathbf{r}$ and a synthesis gate $\beta\in[0,1]$ as:
\begin{equation}
\{\mathbf{r},\ \beta\}=\mathrm{Linear}(\mathrm{Conv}(\mathcal{F},z_{\text{trans}}-z_{t+K-d},z_{0})),
\end{equation} 
The final execution-time latent $\hat{z}_{t+K}$ is reconstructed as $\hat{z}_{t+K} = z_{\text{trans}} + \beta \cdot \mathbf{r}$. The synthesis gate $\beta$ regulates the injection of synthesized features, assigning higher weights to areas where visual dynamics cannot be resolved by spatial transport, as shown in the bottom row of Fig.~\ref{fig: gate_visualize}. In the training phase, we employ an observation prediction loss $\mathcal{L}_{\text{obs}}$ to minimize the discrepancy between the predicted latent $\hat{z}_{t+K}$ and the recorded latent $z_{t+K}$ at execution time as:
\begin{equation}
\mathcal{L}_{\text{obs}}=||z_{t+K} - \hat{z}_{t+K}||_2^2.
\end{equation}
Finally, the VLA policy leverages our anticipated pair to generate the next chunk $\mathbf{A}_{m+1}=\pi_\theta(\cdot\mid\hat{z}_{t+K},\hat{s}_{t+K})$, where the predicted visual latent $\hat{z}_{t+K}$ is directly injected into the policy, bypassing the original vision encoder. The chunk is already aligned to the execution time $t+K$ and is therefore executed directly from its first action, i.e., $\mathbf{A}_{m+1}[0{:}K-1]$.

\subsection{Policy Consistency Loss}
To ensure that the predicted observation $\hat{z}_{t+K}$ and corrected state $\hat{s}_{t+K}$ are faithfully consumed by the policy, we further introduce a policy consistency loss $\mathcal{L}_{\text{policy}}$ to encourage the action chunk generated by the untouched policy from the anticipated pair $(\hat{z}_{t+K},\hat{s}_{t+K})$ to align with the one generated from the ground-truth pair $(z_{t+K},s_{t+K})$. To circumvent the computational overhead of multi-step flow matching during training, we employ a single-step flow approximation to generate the action chunk. The $\mathcal{L}_{\text{policy}}$ is defined as:
\begin{equation}
\mathcal{L}_{\text{policy}}=||\pi_{\theta}(z_{t+K},s_{t+K})-\pi_{\theta}(\hat{z}_{t+K},\hat{s}_{t+K})||_2^2,
\end{equation}
where $\pi_{\theta}(\cdot)$ is the frozen VLA policy. The overall objective function is formulated as $\mathcal{L}_{\text{total}} = \mathcal{L}_{\text{state}} + \mathcal{L}_{\text{obs}} + \lambda \mathcal{L}_{\text{policy}}$.

\section{Experiments}\label{sec: experiments}
In this section, we conduct comprehensive experiments on both simulated and real-world tasks to evaluate the effectiveness of our method under various inference delay settings.

\subtitle{Implementation Details.} We implement the proposed method with PyTorch on NVIDIA RTX 4090 GPUs with the batch size set to 128, reaching convergence within $200\text{k}$ iterations. We employ the Adam optimizer~\cite{Adam} for optimization with the learning rate decayed from $3\times10^{-4}$ to $1\times10^{-6}$, and empirically set the hyperparameter $\lambda=10$. To make our approach robust across a wide range of inference delays without per-delay retraining, we uniformly sample the delay $d\sim\mathcal{U}[1,d_{\max}]$ at each training step, where $d_{\max}$ denotes the largest delay considered on each benchmark. Importantly, the base VLA policy remains untouched throughout the entire training process.
\begin{table*}[!t]
  \centering
  \caption{Performance comparison of $\pi_{0.5}$~\cite{pi05} and SmolVLA-450M~\cite{Smolvla} in the LIBERO~\cite{Libero} benchmark with different methods across various inference delays $d\in \{5,10,15,20\}$. The best results are highlighted in \textbf{bold}. ``Infer.'' and ``Train.'' denote the methods that are categorized as inference-time and training-time, respectively.}
  \resizebox{\textwidth}{!}{
    \begin{tabular}{c|l|cccc|cccc}
    \toprule
    \multirow{3}[6]{*}{} & \multirow{3}[6]{*}{Methods} & \multicolumn{4}{c|}{$\pi_{0.5}$} & \multicolumn{4}{c}{SmolVLA-450M} \\
    \cmidrule{3-10} & & \multicolumn{4}{c|}{Success Rate (\%) $\uparrow$ / Execution Steps $\downarrow$} & \multicolumn{4}{c}{Success Rate (\%) $\uparrow$ / Execution Steps $\downarrow$} \\
    \cmidrule{3-10} & & $d=5$ & $d=10$ & $d=15$ & $d=20$ & $d=5$ & $d=10$ & $d=15$ & $d=20$ \\
    \midrule
    \multirow{4}[2]{*}{\rotatebox{90}{Infer.}} 
    & Naive Async. & 89.5 / 169.0 & 82.7 / 183.0 & 76.3 / 196.6 & 68.3 / 210.9 & 70.3 / 206.4 & 64.5 / 218.2 & 61.3 / 224.2 & 56.2 / 232.8 \\
    & TE    & 87.8 / 173.0 & 84.3 / 180.1 & 80.6 / 185.8 & 74.8 / 197.0 & 68.4 / 208.8 & 66.7 / 212.9 & 62.6 / 220.0 & 59.7 / 224.2 \\
    & BID   & 90.3 / 168.2 & 84.0 / 180.5 & 76.1 / 197.3 & 71.3 / 205.8 & 70.0 / 207.6 & 65.0 / 216.4 & 61.7 / 223.0 & 57.1 / 229.4 \\
    & RTC   & 89.9 / 168.4 & 83.2 / 181.6 & 77.5 / 194.2 & 73.7 / 200.5 & 69.7 / 208.1 & 64.9 / 217.0 & 62.6 / 220.9 & 58.7 / 226.3 \\
    \midrule
    \multirow{4}[2]{*}{\rotatebox{90}{Train.}} 
    & T-RTC & 89.4 / 168.5  &  84.0 / 177.7  &  77.9 / 193.1  &  74.0 / 200.1 & 69.9 / 208.7 & 65.6 / 215.8 & 61.7 / 222.0 & 59.4 / 226.6 \\
    & VLASH & 88.0 / 171.9 & 83.4 / 178.9 & 79.5 / 187.4 & 74.5 / 197.4 & 69.3 / 208.5 & 65.8 / 217.6 & 64.7 / 215.6 & 61.9 / 220.9 \\
    & REMAC & 90.8 / 166.6 & 85.4 / 177.6 & 78.5 / 192.6 & 73.9 / 199.6 & 70.9 / 206.2 & 66.5 / 213.5 & 62.2 / 220.9 & 59.7 / 224.9 \\
    & Ours  & \textbf{94.2} / \textbf{162.4} & \textbf{92.4} / \textbf{166.9} & \textbf{91.0} / \textbf{170.1} & \textbf{88.5} / \textbf{175.1} & \textbf{75.8} / \textbf{198.7} & \textbf{73.3} / \textbf{202.7} & \textbf{71.6} / \textbf{206.4} & \textbf{69.4} / \textbf{209.4} \\
    \bottomrule
    \end{tabular}}
  \label{tab: LIBERO_results}
\end{table*}

\subsection{Simulated Benchmark}
In this section, we conduct experiments on two simulated robotic manipulation benchmarks, including Kinetix~\cite{RTC} and LIBERO~\cite{Libero}, to evaluate the effectiveness of the proposed method under various inference delay settings. We compare our method with two categories of existing methods: (1) Inference-time methods: including \textbf{Naive Asynchronous (Naive Async.)}, which directly executes actions from the most recently generated action chunk, \textbf{Temporal Ensembling (TE)}~\cite{ACT}, \textbf{BID}~\cite{BID}, and \textbf{RTC}~\cite{RTC}. (2) Training-time methods: including \textbf{training-time RTC (T-RTC)}~\cite{training_rtc}, \textbf{VLASH}~\cite{VLASH}, \textbf{REMAC}~\cite{REMAC}, and \textbf{DiscreteRTC}~\cite{DiscreteRTC}. 

\subtitle{Kinetix.} We first evaluate our method on the Kinetix simulator~\cite{RTC}, which contains 12 highly dynamic and stochastic environments designed to assess asynchronous execution. For each task, we adopt the action chunking flow policies with a prediction horizon of $H=8$, pretrained for 32 epochs following~\cite{RTC}. The inference delay $d$ is varied from 0 to 4, and the execution horizon is selected from $\max\{1,d\}$ to $8-d$, ensuring that actions remain continuously available without gaps between consecutive chunks.
\begin{figure}[!t]
    \centering
    \includegraphics[width=\linewidth]{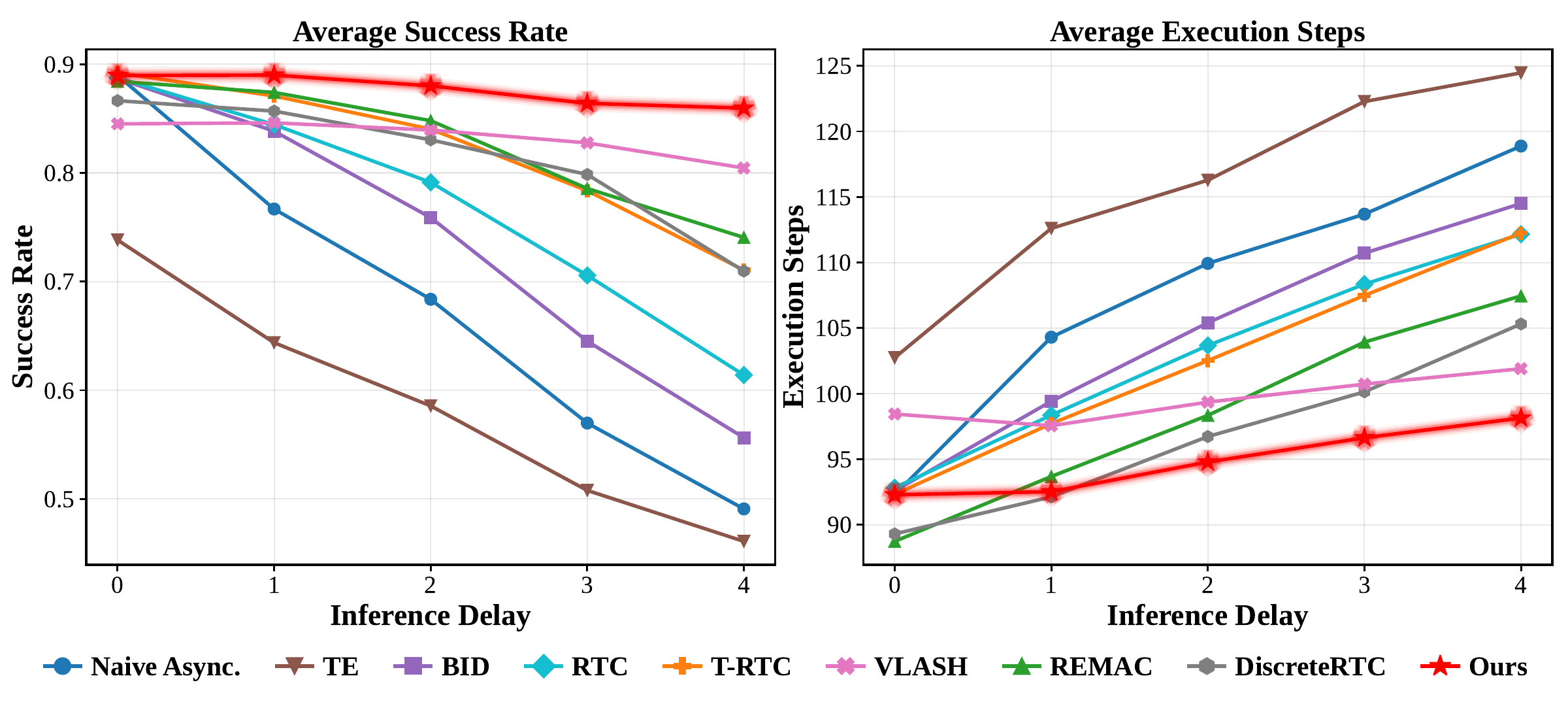}
    \caption{Performance comparison in the Kinetix simulator~\cite{RTC} with different execution methods across various inference delays $d\in[0,4]$.}
    \label{fig: Kinetix_results}
\end{figure}

We report the average performance across 12 tasks, measured by both task success rate and execution steps, as shown in Fig.~\ref{fig: Kinetix_results}. Naive Async. suffers performance collapse as the inference delay increases, while inference-time methods provide partial mitigation. Training-time methods show stronger robustness, but their reliance on outdated observations and states still causes notable degradation under large delays. VLASH reduces this misalignment by conditioning on execution-time states, yet its performance remains inferior to ours. In contrast, our method shows superior robustness, with only a 3.0\% success rate drop from $d=0$ to $d=4$, and fewer steps are required when $d\geq2$. Meanwhile, previous methods require more execution steps under large delays due to reactive hesitation and inter-chunk discontinuities, proving that our method can realize reliable and smoother decision-making. It is worth noting that our method is slightly behind several training-time methods that fine-tune or retrain the policy for the benchmark tasks at small delays, i.e., $d=\{0,1\}$, whereas our method keeps the VLA policy frozen and only serves as a plug-and-play module.

\subtitle{LIBERO.} We further evaluate on the LIBERO~\cite{Libero} benchmark, which spans four task suites, including Spatial, Object, Goal, and Long. To assess whether our adapter transfers across backbones, we attach it to two pretrained VLAs, $\pi_{0.5}$~\cite{pi05} and SmolVLA-450M~\cite{Smolvla}, with the execution horizon set to $K=25$. We probe a wider delay as $d\in\{5,10,15,20\}$, reflecting practical deployment where remote or resource-constrained inference on large VLAs incurs substantial delays. For inference-time methods and our method, we build directly on the finetuned weights of the above two VLAs released by LeRobot~\cite{LeRobot} and keep them frozen for evaluation. For training-time methods, instead of adapting these weights, we reproduce them ourselves following their corresponding publications, except for VLASH, which provides an official LIBERO implementation. We omit DiscreteRTC for comparison since it presumes a discrete diffusion policy trained from scratch and cannot be instantiated on the flow-based backbones adopted here.

We report the average success rate and execution steps across four tasks in Table~\ref{tab: LIBERO_results}, where our method consistently achieves the highest success rate while requiring the fewest execution steps on all evaluated delays and both VLA backbones. The performance of existing methods deteriorates substantially with increasing delay, whereas our approach exhibits only a modest 5.7\% and 6.4\% decrease in success rate from $d=5$ to $d=20$ on $\pi_{0.5}$ and SmolVLA, respectively. Moreover, our method consistently completes tasks in fewer execution steps, indicating smoother and more decisive control under severe asynchronous execution. VLASH, which partially mitigates prediction-execution misalignment by rolling forward the proprioceptive state, presents robustness at large delays compared to other methods. However, it still trails our method, suggesting that compensating for the robot state alone is insufficient when visual observations remain temporally stale. By jointly anticipating the execution-time context, our method consistently achieves the best performance. \textbf{We provide more comprehensive simulated experimental analysis in the appendix.}

\subsection{Ablation and Analysis}
We conduct a series of ablations to evaluate our newly proposed components. The average success rate of the LIBERO benchmark using SmolVLA-450M is reported in Table~\ref{tab: ablation}. 

We first construct a Baseline that feeds the stale pair $(o_{t+K-d},s_{t+K-d})$ directly into the policy and executes the resulting chunk from index $[0{:}K-1]$. Without any compensation, as reported in row 1 of Table~\ref{tab: ablation}, the success rate deteriorates rapidly as the delay increases, demonstrating the severe impact of prediction-execution misalignment. Introducing only our State Correction Module (SCM), i.e., row 2 of Table~\ref{tab: ablation}, which replaces the stale state with the corrected state $\hat{s}_{t+K}$ while retaining the stale observation, improves the low-delay results but remains ineffective under large delays, proving that stale visual observation, not only the proprioceptive state, is also the bottleneck under asynchronous execution. In contrast, further introducing the Observation Prediction Module (OPM), i.e., row 3 of Table~\ref{tab: ablation}, which supplies the execution-time visual latent representation $\hat{z}_{t+K}$, provides substantial improvements. Moreover, we additionally employ the policy consistency loss $\mathcal{L}_{\text{policy}}$ to align the predicted observation-state pair with the distribution expected by the VLA policy, thus introducing overall performance improvements as reported in row 4 of Table~\ref{tab: ablation}.

Finally, the last two columns report the computational overhead of the proposed adapter, which introduces only 5.19 M additional parameters and 3.04 ms of inference latency, indicating that our method achieves substantial performance improvements with limited computational overhead.
\begin{table}[!t]
  \centering
  \caption{The average success rate of ablation studies using SmolVLA-450M~\cite{Smolvla} on the LIBERO~\cite{Libero} benchmark. The last two columns report the added parameters (M) and inference time (ms) of our newly proposed modules relative to the VLA policy.}
  \resizebox{\linewidth}{!}{
    \begin{tabular}{l|cccc|cc}
    \toprule
    \multirow{2}[4]{*}{Methods} & \multicolumn{4}{c|}{Success Rate (\%)} & \multirow{2}[4]{*}{Param. } & \multirow{2}[4]{*}{Time} \\
    \cmidrule{2-5} & $d=5$ & $d=10$ & $d=15$ & $d=20$ & &  \\
    \midrule
    Baseline & 26.1 & 8.4 & 2.5 & 0.7 & 450.0 & 35.4 \\
    +SCM & 45.9 & 18.1 & 10.4 & 5.8 & +0.04  & +1.15 \\
    +OPM & 74.1 & 70.4 & 70.0 & 67.5 & +5.19  & +3.04 \\
    + $\mathcal{L}_{\text{policy}}$ (Ours) & 75.8 &  73.3 & 71.6 & 69.4 & - & - \\
    \bottomrule
    \end{tabular}}
  \label{tab: ablation} 
\end{table}

\subsection{Real-World Evaluation}
\subtitle{Setup.} We deploy FutureRTC on a dual-arm AgileX Cobot Magic robot using $\pi_{0.5}$ as the VLA backbone with an execution horizon of $K=25$. The robot operates at a control frequency of $\text{30Hz}$, corresponding to the control period of $\Delta t \approx \text{33ms}$. Without any test-time strategy, a single forward pass of the VLA requires approximately $\text{99ms}$. Since the policy is served on a remote machine, an additional $\sim\text{40ms}$ is incurred by LAN communication, while data processing and disk logging contribute another $\sim\text{30ms}$. The resulting end-to-end latency is $\sim\text{170ms}$, corresponding to an inference delay of about $d=5$. To emulate more challenging remote or resource-constrained deployment scenes, we further inject an additional $\text{150ms}$ latency, extending the delay to $d=10$.

\subtitle{Task Design and Measurement.} We evaluate our method and competitive methods, including Naive Async., RTC, and VLASH, on three bimanual manipulation tasks: (1) \emph{Stack Plates}, stacking the left and right plates onto the middle one; (2) \emph{Fold Towel}, folding a towel with both arms to form a compact square; and (3) \emph{Hang Cups}, hanging the left and right cups onto the corresponding sides of a rack, respectively. For each task, we collect 100 demonstrations to fine-tune $\pi_{0.5}$. Evaluations are conducted across 20 trials per task, in which the initial position and orientation of the object are randomized. The maximum execution steps are strictly limited to 400, 450, and 400 for the three tasks, respectively.
\begin{figure}[!t]
    \centering
    \includegraphics[width=\linewidth]{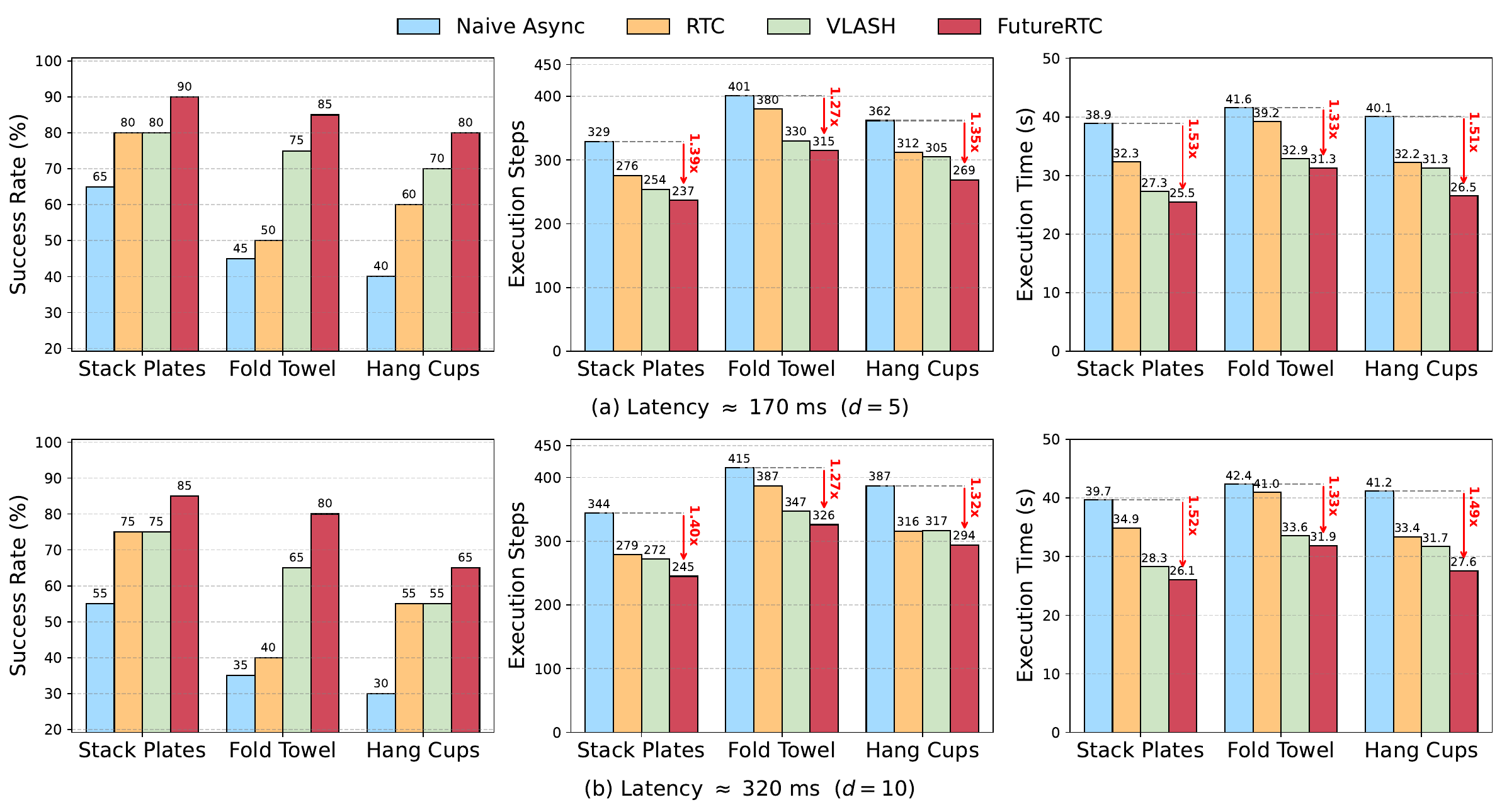}
    \caption{Real-world comparison on three bimanual manipulation tasks, \emph{Stack Plates}, \emph{Fold Towel}, and \emph{Hang Cups}, with $\pi_{0.5}$~\cite{pi05} as the VLA policy.}
    \label{fig: realworld_results}
\end{figure}

As illustrated in Fig.~\ref{fig: realworld_results}, our FutureRTC consistently achieves the highest success rates, as well as the fewest execution steps and times across three tasks, particularly as latency increased. While Naive Async. degrades sharply, RTC and VLASH offer only partial relief from prediction-execution misalignment, ultimately falling short of FutureRTC in terms of effectiveness and efficiency. With the ability to predict execution-time context, FutureRTC maintains reliable and smooth operation even in the presence of remarkable inference latency, indicating that the robustness observed in simulations has effectively transferred to real-world operations. \textbf{We provide more comprehensive real-world experimental analysis in the appendix.}

\section{Conclusion}\label{sec: conclusion}
We have presented FutureRTC, a plug-and-play adaptation framework for mitigating the prediction-execution misalignment inherent in asynchronous VLA execution. Specifically, FutureRTC anticipates the execution-time context by correcting rolled-forward proprioceptive states and predicting future visual representations through the proposed state correction module and observation prediction module, respectively, enabling the VLA policy to generate actions conditioned on the observation and state at execution time. Moreover, the policy consistency loss is introduced to encourage the predicted context to be faithfully consumed by the policy. Extensive experiments on simulated benchmarks and real-world manipulation tasks demonstrate that FutureRTC consistently improves delay robustness, execution smoothness, task efficiency, and success rates across VLA backbones.

\subtitle{Acknowledgements.} This work was supported by the National Natural Science Foundation of China (NSFC) under grants No.625B2123, No.62372091, and No.52502425, and the Hainan Province Science and Technology Plan Project under grant ZDYF2024(LALH)001.

\bibliography{aaai2027}

\clearpage

\appendixpage
This appendix is organized as follows:
\begin{itemize}
   \item \textbf{Sec. More Detailed Analysis} formally defines real-time robot execution, analyzes the impact of the prediction-execution misalignment, and reports the computational efficiency and trajectory smoothness of our method.
   \item \textbf{Sec. Real-world Evaluation} details the video comparisons, hardware setup, task design, and policy implementation of our real-world experiments.
   \item \textbf{Sec. Simulated Evaluation} presents the per-task results on Kinetix and the detailed per-suite results on LIBERO.
   \item \textbf{Sec. Ablation Study} reports the complete ablation of our proposed components on both SmolVLA-450M and $\pi_{0.5}$.
   \item \textbf{Sec. Limitations} discusses the limitations and future directions of our method.
\end{itemize}


\section{More Detailed Analysis}\label{sec: more_details}
\subsection{Definition of Real-Time Robot Execution}\label{subsec: def_realtime}
VLA policies are typically trained on demonstrations under the idealized assumption of instantaneous observation acquisition and immediate action execution. During deployment, however, generating an action chunk incurs a non-negligible inference latency $\delta$, compounded by hardware-level communication, readout, and actuation delays. Following~\cite{real_time_vla_v2}, we define real-time robot execution as the continuous, stall-free operation of the VLA policy at a target control frequency, satisfying three core criteria:
\begin{itemize}
    \item \textbf{Accurate.} The policy reliably completes the intended task with a high success rate.
    \item \textbf{Smooth.} The commanded trajectory is smooth, free of abrupt accelerations and jerky motions.
    \item \textbf{Fast.} The policy completes the task in fewer control steps and less wall-clock time.
\end{itemize}

Satisfying the above three criteria exposes a fundamental tension between synchronous and asynchronous execution. Synchronous execution conditions the policy on the latest observation and proprioceptive state, thereby preserving action accuracy, but forces the robot to idle during every chunk inference, producing stalls that directly violate the fast criterion. Asynchronous execution eliminates stalls by generating the next action chunk in parallel with the execution of the current one. However, each chunk is conditioned on observations and states that have already become stale by the time it is executed, giving rise to the \emph{prediction-execution misalignment}. Thus, the generated actions become progressively misaligned with the evolving scene, leading to inter-chunk discontinuities and simultaneously degrading accuracy, smoothness, and execution efficiency as the inference delay increases.

Overall, real-time control requires the stall-free property of asynchronous execution without sacrificing execution-time alignment that our method aims to address.
\begin{table*}[!t]
  \centering
  \caption{Analysis of prediction-execution misalignment on the LIBERO~\cite{Libero} benchmark, measured by average success rates across four suites, with $\pi_{0.5}$~\cite{pi05} and SmolVLA-450M~\cite{Smolvla} as the VLA policies.}
  \resizebox{\textwidth}{!}{
    \begin{tabular}{ll|cccc|cccc}
    \toprule
    \multirow{2}{*}{Backbone} & \multirow{2}{*}{Input} & \multicolumn{4}{c|}{Execute $\mathbf{A}[0:K-1]$} & \multicolumn{4}{c}{Execute $\mathbf{A}[d:d+K-1]$} \\
    \cmidrule{3-10}
    & & $d=5$ & $d=10$ & $d=15$ & $d=20$ & $d=5$ & $d=10$ & $d=15$ & $d=20$ \\
    \midrule
    \multirow{5}{*}{$\pi_{0.5}$}
    & $\{o_{t+K-d},\, s_{t+K-d}\}$ & 46.9 & 11.7 & 5.2 & 2.5 & 89.5 & 82.7 & 76.3 & 68.3 \\
    & $\{o_{t+K-d},\, s_{t+K}\}$   & 47.2 & 12.4 & 5.3 & 1.9 & 89.7 & 82.6 & 73.4 & 67.6 \\
    & $\{o_{t+K},\, s_{t+K-d}\}$   & 96.3 & 95.9 & 95.4 & 95.8 & 37.6 & 4.7 & 1.3 & 0.8 \\
    & $\{o_{t+K},\, s_{t+K}\}$     & 96.6 & 96.6 & 96.6 & 96.6 & 37.2 & 6.5 & 1.6 & 0.9 \\
    & \textcolor{gray}{$\{\hat{z}_{t+K},\, \hat{s}_{t+K}\}$ (Ours)} & \textcolor{gray}{94.2} & \textcolor{gray}{92.4} & \textcolor{gray}{91.0} & \textcolor{gray}{88.5} & \textcolor{gray}{39.0} & \textcolor{gray}{7.2} & \textcolor{gray}{2.6} & \textcolor{gray}{0.7} \\
    \midrule
    \multirow{5}{*}{SmolVLA}
    & $\{o_{t+K-d},\, s_{t+K-d}\}$ & 26.1 & 8.4 & 2.5 & 0.7 & 70.3 & 64.5 & 61.3 & 56.2 \\
    & $\{o_{t+K-d},\, s_{t+K}\}$   & 44.6 & 17.9 & 10.9 & 6.6 & 59.7 & 31.5 & 22.1 & 15.6 \\
    & $\{o_{t+K},\, s_{t+K-d}\}$   & 61.7 & 26.8 & 18.1 & 16.4 & 39.3 & 7.5 & 2.3 & 1.0 \\
    & $\{o_{t+K},\, s_{t+K}\}$     & 77.6 & 77.6 & 77.6 & 77.6 & 24.0 & 4.0 & 1.9 & 0.3 \\
    & \textcolor{gray}{$\{\hat{z}_{t+K},\, \hat{s}_{t+K}\}$ (Ours)} & \textcolor{gray}{75.8} & \textcolor{gray}{73.3} & \textcolor{gray}{71.6} & \textcolor{gray}{69.4} & \textcolor{gray}{24.2} & \textcolor{gray}{3.9} & \textcolor{gray}{0.7} & \textcolor{gray}{0.2} \\
    \bottomrule
    \end{tabular}}
  \label{tab: misalignment}
\end{table*}

\subsection{Impact of the Prediction-Execution Misalignment}
To quantify the individual impact of observation and proprioceptive state staleness under asynchronous execution, we conduct experiments on the LIBERO~\cite{Libero} benchmark with $\pi_{0.5}$~\cite{pi05} and SmolVLA-450M~\cite{Smolvla}. At the generation time $t+K-d$, the policy is conditioned on the stale pair $(o_{t+K-d}, s_{t+K-d})$, while the generated chunk is executed at $t+K$. We isolate the contribution of each modality by selectively replacing stale inputs with their execution-time counterparts, i.e., the ground-truth observation $o_{t+K}$ or state $s_{t+K}$, resulting in four configurations as follows:
\begin{itemize}
    \item \textbf{Stale observation + stale state}, i.e., $\{o_{t+K-d}, s_{t+K-d}\}$, corresponding to naive asynchronous execution.
    \item \textbf{Stale observation + execution-time state}, i.e., $\{o_{t+K-d}, s_{t+K}\}$, correcting only the state.
    \item \textbf{Execution-time observation + stale state}, i.e., $\{o_{t+K}, s_{t+K-d}\}$, correcting only the visual observation.
    \item \textbf{Execution-time observation + execution-time state}, i.e., $\{o_{t+K}, s_{t+K}\}$, serving as the delay-free upper bound.
\end{itemize}

Moreover, since the appropriate execution of an action chunk $\mathbf{A}$ depends on whether its conditioning inputs are aligned with the execution time, we further clarify the chunk-splicing conventions used in the main paper. $\mathbf{A}[0{:}K{-}1]$ executes the generated chunk from its first action, which is valid when the conditioning context corresponds to the execution time. In contrast, $\mathbf{A}[d{:}d{+}K{-}1]$ removes the first $d$ actions that overlap with the already committed chunk and executes the remaining actions, following the naive asynchronous convention. We evaluate the above four configurations under both execution conventions to disentangle the impact of stale conditioning inputs from that of chunk-splicing offsets.
 
The average success rate across four suites of the LIBERO benchmark is reported in Table~\ref{tab: misalignment}. We first examine how the chunk-splicing convention interacts with the staleness of the conditioning inputs. Under naive asynchronous execution, the policy is conditioned on the fully stale pair $\{o_{t+K-d}, s_{t+K-d}\}$. Executing the generated chunk directly from its first action, i.e., $\mathbf{A}[0{:}K{-}1]$, causes the success rate to collapse rapidly as the delay increases. The reason is that the chunk is generated for the time index $t+K-d$ but executed $d$ steps later. Discarding the first $d$ actions, i.e., executing $\mathbf{A}[d{:}d{+}K{-}1]$, compensates for this temporal offset and substantially improves performance. However, discarding the first $d$ actions is beneficial only when the conditioning inputs are stale. Once the execution-time context becomes available, i.e., the $\{o_{t+K}, s_{t+K-d}\}$ and $\{o_{t+K}, s_{t+K}\}$, the chunk is generated as if the current step is at $t+K$, so its first action already corresponds to the execution time. Removing the first $d$ actions then skips into the future, commanding actions intended for $t+K+d$ before the robot has reached the corresponding state. Consequently, $\mathbf{A}[d{:}d{+}K{-}1]$ collapses, whereas $\mathbf{A}[0{:}K{-}1]$ stays near the upper bound. Hence, whenever the execution-time context can be recovered, the chunk should be executed directly from index $[0{:}K{-}1]$.

Under the correct execution convention $\mathbf{A}[0{:}K{-}1]$, we further compare the four conditioning configurations. Correcting only the state, i.e., ${o_{t+K-d}, s_{t+K}}$, yields little improvement over the fully stale baseline on $\pi_{0.5}$, which is consistent with its proprioceptive state being encoded as textual tokens within the language input. In contrast, state correction brings a more noticeable gain on SmolVLA, whose continuous state is projected through a dedicated linear layer into a separate state token directly attended by the action expert. Moreover, correcting only the observation, i.e., $\{o_{t+K}, s_{t+K-d}\}$, recovers the majority of the performance gap. These results demonstrate that the stale visual observation, not only the stale state, is also a dominant source of prediction-execution misalignment.
\begin{table}[!t]
  \centering
  \caption{The parameters and inference time introduced by our method to $\pi_{0.5}$~\cite{pi05} and SmolVLA-450M~\cite{Smolvla}.}
  \resizebox{\linewidth}{!}{
  \begin{tabular}{l|cc|cc}
    \toprule
    \multirow{2}[4]{*}{Methods} & \multicolumn{2}{c|}{$\pi_{0.5}$} & \multicolumn{2}{c}{SmolVLA} \\
    \cmidrule{2-5} & \multicolumn{1}{l}{Param. (M)} & \multicolumn{1}{l|}{Time (ms)} & \multicolumn{1}{l}{Param. (M)} & \multicolumn{1}{l}{Time (ms)} \\
    \midrule
    Base & 3616.8 & 98.8 & 450.0 & 35.4 \\
    \midrule
    +FutureRTC  & +6.45 & +3.64 & +5.19 & +3.04 \\
    \bottomrule
    \end{tabular}}
  \label{tab: computational}
\end{table}

Finally, when both the execution-time observation and state are supplied, i.e., $\{o_{t+K}, s_{t+K}\}$ with $\mathbf{A}[0{:}K{-}1]$, the success rate remains constant across all delays, matching the delay-free performance. This result indicates that asynchronous execution itself incurs no performance penalty, provided that the policy is conditioned on the correct execution-time context. Since such context is unavailable at generation time, FutureRTC instead predicts the execution-time observation $\hat{z}_{t+K}$ and state $\hat{s}_{t+K}$ from their stale counterparts, enabling the policy to operate as if the delay did not exist and thereby approach this delay-invariant upper bound.
\begin{figure}[!t]
    \centering
    \includegraphics[width=\linewidth]{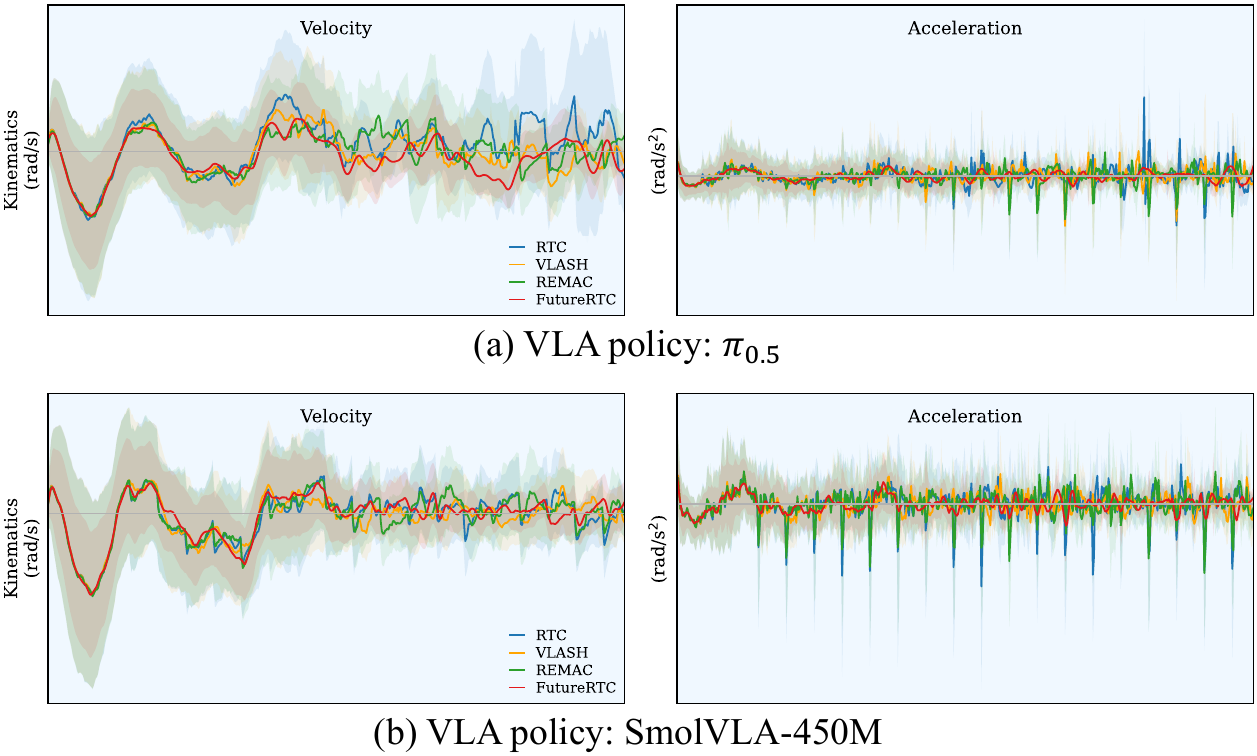}
    \caption{Average action velocity and acceleration on the LIBERO~\cite{Libero} benchmark under $d=5$ for $\pi_{0.5}$~\cite{pi05} (a) and SmolVLA-450M~\cite{Smolvla} (b), comparing RTC~\cite{RTC}, VLASH~\cite{VLASH}, REMAC~\cite{REMAC}, and our FutureRTC. The left panel plots the action velocity ($\mathrm{rad/s}$) and the right panel the action acceleration ($\mathrm{rad/s^2}$) during execution, where previous methods exhibit pronounced acceleration spikes at the chunk boundaries, whereas FutureRTC produces smoother trajectories.}
    \label{fig: kinematics}
\end{figure}

\subsection{Computational Efficiency}
We evaluated the computational efficiency of our method in terms of the parameters and inference time introduced to the base VLA policies $\pi_{0.5}$~\cite{pi05} and SmolVLA-450M~\cite{Smolvla}. As reported in Table~\ref {tab: computational}, our method introduces only marginal overhead on both VLA policies, adding 6.45M parameters and 3.64ms inference latency to $\pi_{0.5}$, and 5.19M parameters and 3.04ms to SmolVLA, showing that FutureRTC achieves robust real-time execution at limited computational cost.

\subsection{Trajectory Smoothness Analysis}
To evaluate the smoothness criterion, we analyze the kinematic properties of the executed actions on the LIBERO~\cite{Libero} benchmark across all four suites, comparing FutureRTC with representative baselines including RTC~\cite{RTC}, VLASH~\cite{VLASH}, and REMAC~\cite{REMAC}. Fig.~\ref{fig: kinematics} reports the average action velocity and acceleration profiles during execution for each backbone, where previous methods suffer from pronounced spikes caused by inter-chunk discontinuities. In contrast, VLASH partially mitigates the spikes by conditioning on the forward-predicted future state, yet it remains less smooth than our approach owing to the lack of execution-time observation as input for action prediction. By anticipating the execution-time context, our FutureRTC consistently produces the smoothest trajectories.

\section{Real-world Evaluation}\label{sec: more_real_world}
To qualitatively demonstrate the real-world execution behavior of FutureRTC, we provide side-by-side comparison videos in the \href{https://jianghaiscu.github.io/FutureRTC_proj/}{Project Website}. For each of the three real-world bimanual tasks, i.e., \emph{Stack Plates}, \emph{Fold Towel}, and \emph{Hang Cups}, we visualize executions of Naive Async., RTC~\cite{RTC}, VLASH~\cite{VLASH}, and FutureRTC under asynchronous inference. The videos show that FutureRTC generates smoother and more decisive motions with fewer hesitations, while achieving more reliable task completion, whereas the baselines exhibit jerky transitions and delayed completion.

\subsection{Hardware Setup}\label{subsec: rw_setup}
\begin{figure}[!t]
    \centering
    \includegraphics[width=\linewidth]{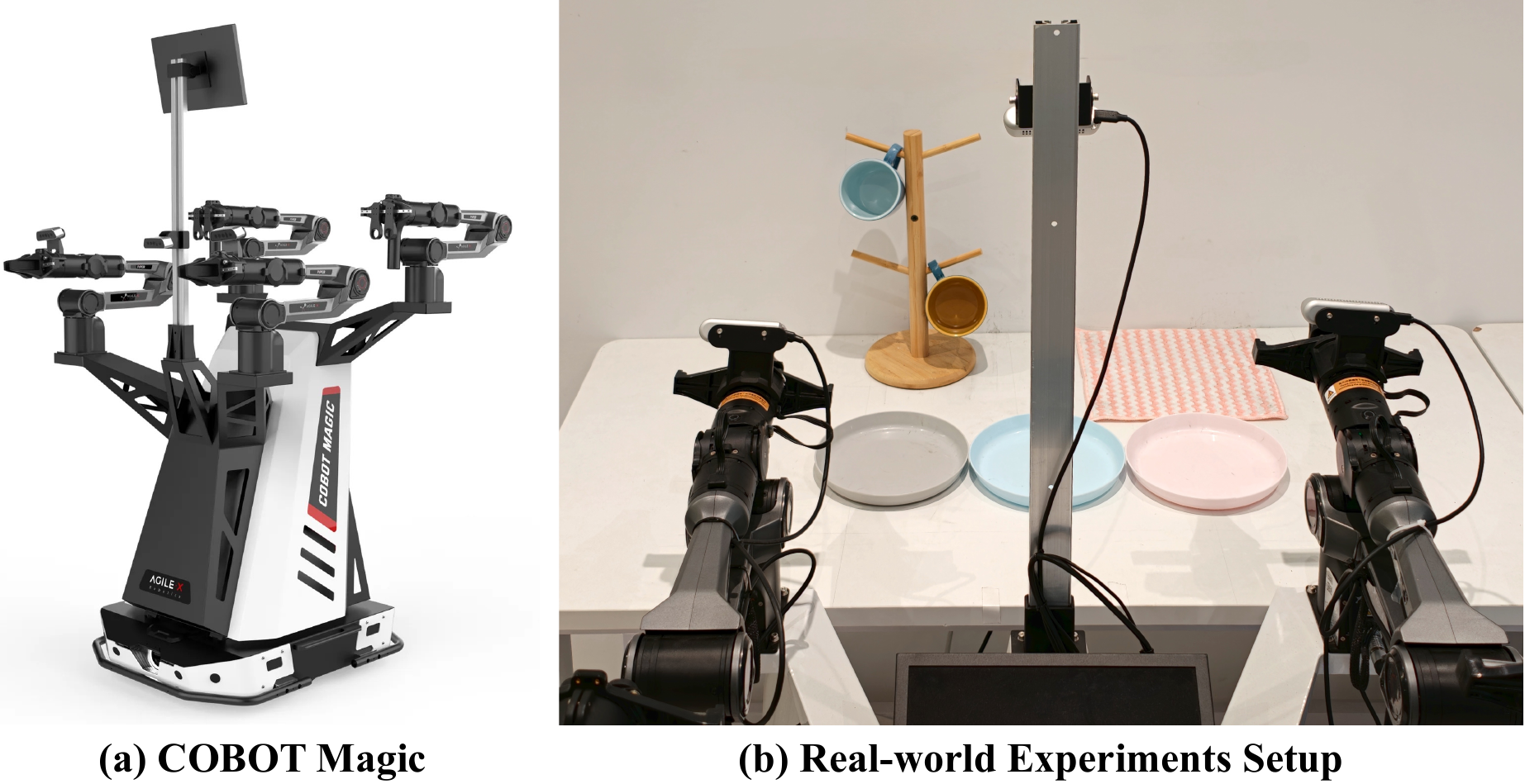}
    \caption{Real-world experimental platform. \textbf{(a)} The AgileX Cobot Magic robot. \textbf{(b)} The workspace of the three bimanual tasks, containing the plates, towel, and cups with a cup rack.}
    \label{fig: setup}
\end{figure}

We conduct real-world experiments on a dual-arm AgileX Cobot Magic robot\footnote{\url{https://global.agilex.ai/products/cobot-magic}} equipped with three RGB cameras, i.e., two wrist-mounted cameras and one global camera mounted on the central mast, as shown in Fig.~\ref{fig: setup} (a). We also illustrate the objects used in three tasks in Fig.~\ref{fig: setup} (b), i.e., the plates, towel, and cups with a cup rack used in \emph{Stack Plates}, \emph{Fold Towel}, and \emph{Hang Cups}, respectively.
\begin{figure}[!t]
    \centering
    \includegraphics[width=\linewidth]{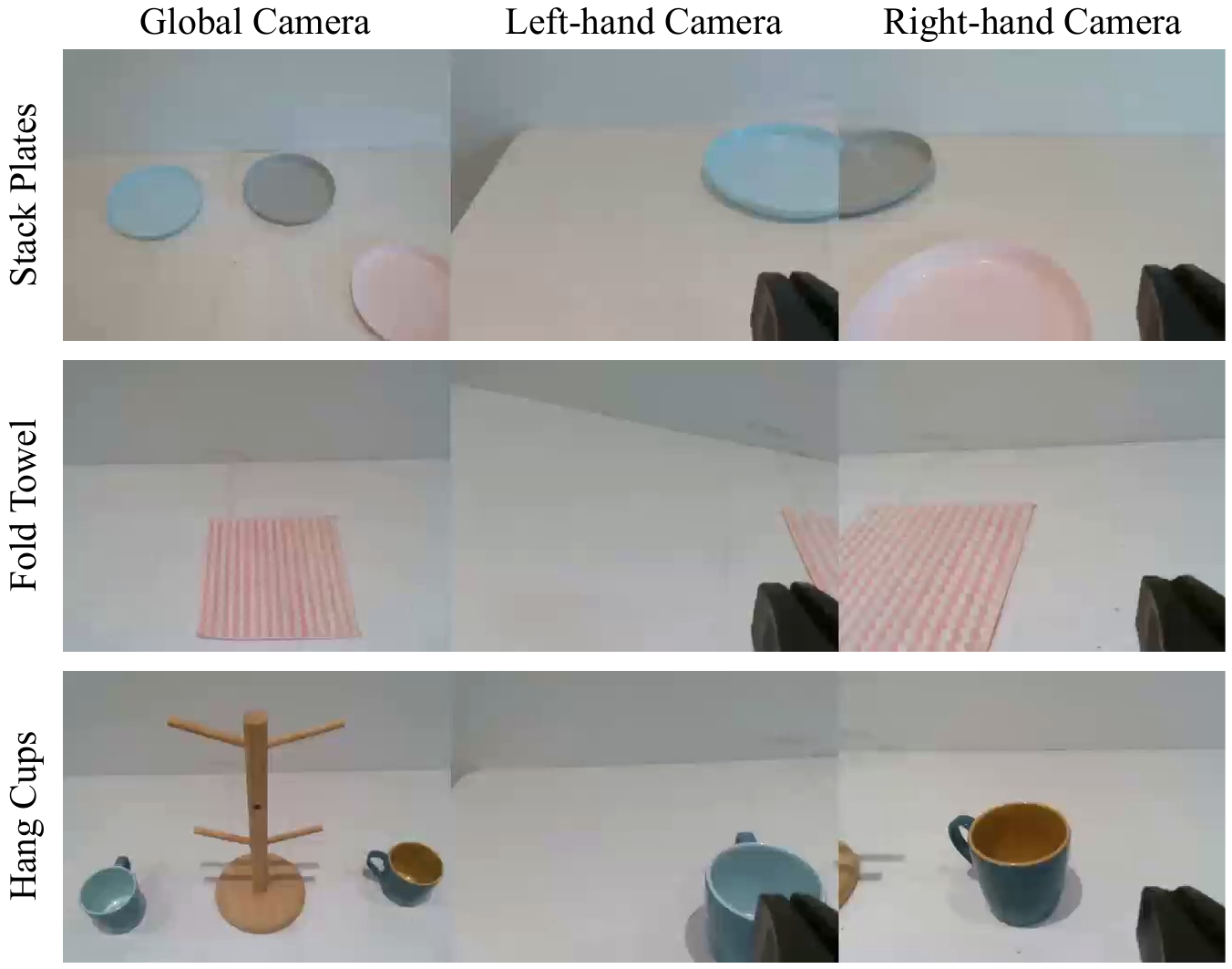}
    \caption{Example demonstration sequences of the three real-world tasks: \emph{Stack Plates}, \emph{Fold Towel}, and \emph{Hang Cups}.}
    \label{fig: tasks}
\end{figure}

\subsection{Task Design}\label{subsec: rw_tasks}
We design three bimanual manipulation tasks to evaluate coordinated two-arm control under asynchronous execution:
\begin{itemize}
    \item \textbf{Stack Plates}: the two arms grasp the left and right plates and stack them onto the middle plate to form a single pile.
    \item \textbf{Fold Towel}: the two arms fold a flattened towel with two successive folds into a compact square. 
    \item \textbf{Hang Cups}: the two arms pick up the left and right cups and hang them onto the corresponding branches of a rack.
\end{itemize}

For each task, we collect 100 teleoperated demonstrations for policy fine-tuning, with the initial object positions and orientations randomized across demonstrations. We illustrate the demonstration data of the three tasks in Fig.~\ref{fig: tasks}.

\subsection{Policy Implementation}\label{subsec: rw_finetune}
To obtain a task-competent policy, we fine-tune $\pi_{0.5}$~\cite{pi05} on the collected demonstrations of each task for 30k iterations. This fine-tuned policy is kept frozen and shared across the inference-time methods, including Naive Async. and RTC, and our method, consistent with the plug-and-play setting used throughout the main paper. The fine-tuning hyperparameters are summarized in Table~\ref{tab: rw_finetune}. For the training-based method VLASH, we adopt the officially released implementation to fine-tune it on $\pi_{0.5}$.

\section{Simulated Evaluation}\label{sec: more_simulated}
\begin{figure*}[!t]
    \centering
    \includegraphics[width=\linewidth]{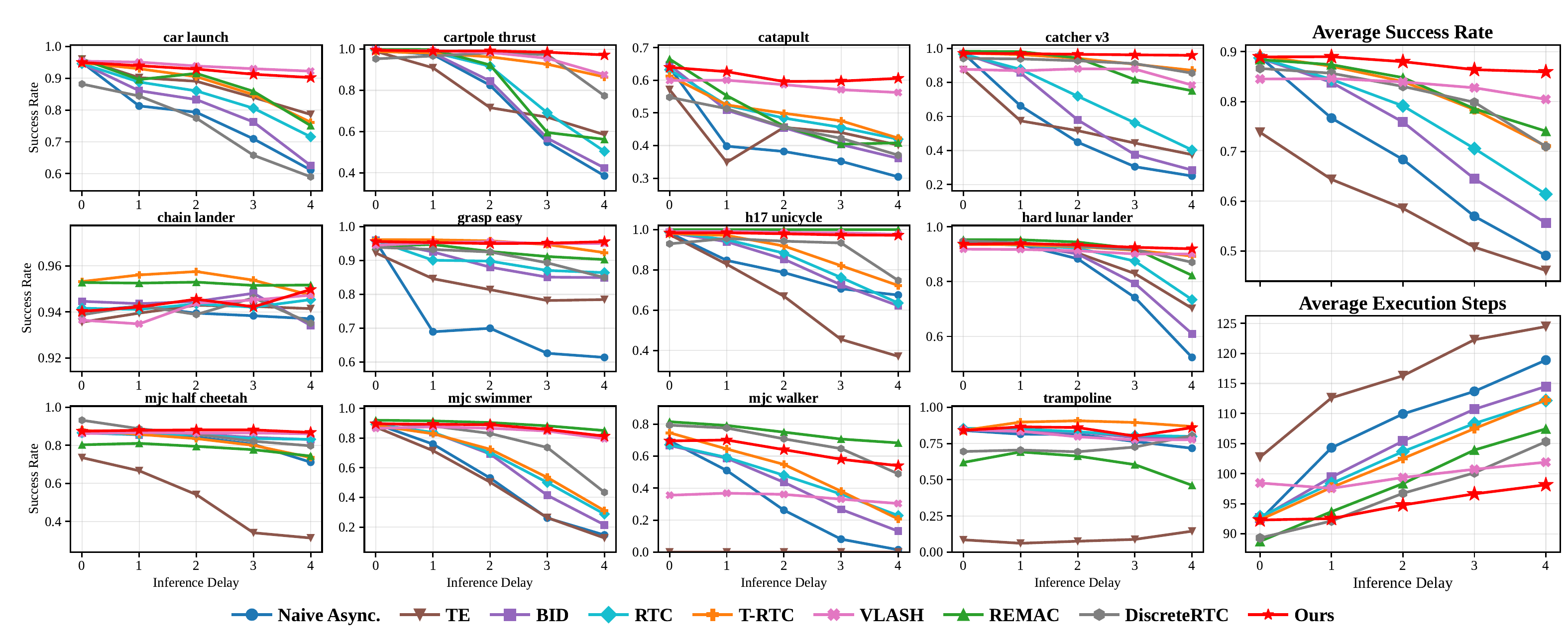}
    \caption{Performance comparison in the Kinetix simulator~\cite{RTC} with different methods across various inference delays $d\in[0,4]$. \textbf{Left:} Success rate for individual tasks. \textbf{Right:} Average success rate and execution steps.}
    \label{fig: kinetix_supp}
\end{figure*}

\subtitle{Kinetix.} We report the per-task success rates across all 12 Kinetix environments under inference delays $d\in[0,4]$ in Fig.~\ref{fig: kinetix_supp}. FutureRTC demonstrates consistently robust performance across diverse environments, exhibiting the smallest and most consistent degradation as the delay increases, without the abrupt failures observed in other comparison methods on individual tasks. Although FutureRTC does not dominate every individual environment, its consistent robustness across diverse dynamics yields the strongest overall performance under varying inference delays.

\subtitle{LIBERO.} We report the per-suite success rate (SR) and execution steps on the four LIBERO suites in Tables~\ref{tab: libero_pi05} and~\ref{tab: libero_smolvla} for $\pi_{0.5}$ and SmolVLA-450M, respectively, complementing the averaged results in the main paper. We additionally include the synchronous (Sync.) execution results as reference, i.e., $d=0$. FutureRTC consistently achieves the highest success rate and the fewest execution steps across all suites and delay settings, with the largest performance gains observed on the long-horizon suite, i.e., Long, where prediction-execution misalignment accumulates most severely. 
\begin{table}[!t]
  \centering
  \caption{The hyperparameters adopted for fine-tuning $\pi_{0.5}$ on the real-world demonstrations.}
  \resizebox{\linewidth}{!}{
  \begin{tabular}{l|c}
    \toprule
    Hyperparameter & Value \\
    \midrule
    Base policy                & $\pi_{0.5}$\\
    Fine-tuning scheme         & Full fine-tuning \\
    Optimizer                  & AdamW ($\beta_1{=}0.9$, $\beta_2{=}0.95$, $\epsilon{=}10^{-8}$) \\
    Weight decay               & $1\times10^{-10}$ \\
    LR schedule                & Cosine, $1{,}000$ warmup steps \\
    Peak / final learning rate & $2.5\times10^{-5}$ / $2.5\times10^{-6}$ \\
    Training steps             & $30{,}000$\\
    Batch size                 & $8$ \\
    Action horizon ($H$)       & $50$ \\
    Random seed                & $42$ \\
    \# Demonstrations per task  & $100$ \\
    Training hardware          & $4{\times}$ NVIDIA RTX 4090 24GB \\
    \bottomrule
  \end{tabular}}
  \label{tab: rw_finetune}
\end{table}

\section{Ablation Study}\label{sec: more_ablation}
We report the detailed ablation results of $\pi_{0.5}$~\cite{pi05} and SmolVLA-450M~\cite{Smolvla} on the LIBERO~\cite{Libero} benchmark across four suites and all delays $d\in\{5,10,15,20\}$ in Table~\ref{tab: ablation_supp}. We first construct a Baseline that feeds the stale pair $(o_{t+K-d}, s_{t+K-d})$ directly into the policy and executes the resulting chunk from index $[0{:}K{-}1]$, in which the success rate deteriorates rapidly as the delay increases on each suite and both backbones, demonstrating the severe impact of the prediction-execution misalignment. Introducing only the State Correction Module (SCM), which replaces the stale state with the corrected state $\hat{s}_{t+K}$ while retaining the stale observation, yields limited improvement that stays far from recovering the performance under large delays, proving that the stale visual observation is also a critical bottleneck under asynchronous execution. Further introducing the Observation Prediction Module (OPM), which supplies the execution-time visual latent representation $\hat{z}_{t+K}$, provides substantial improvements across all suites and delays. Finally, we employ the policy consistency loss $\mathcal{L}_{\text{policy}}$ to align the predicted observation-state pair with the distribution expected by the policy, bringing a further consistent performance gain. 

\begin{table*}[!t]
  \centering
  \caption{Detailed performance comparison of $\pi_{0.5}$~\cite{pi05} in the LIBERO~\cite{Libero} benchmark with different methods across various inference delays $d\in \{5,10,15,20\}$. ``Infer.'' and ``Train.'' denote the methods that are categorized as inference-time and training-time, respectively. ``SR'' and ``Steps'' indicate success rate and execution steps, respectively.}
  \resizebox{\textwidth}{!}{
    \begin{tabular}{c|c|c|cccccccc|cc}
    \toprule
    \multirow{2}[4]{*}{} & \multirow{2}[4]{*}{Methods} & \multirow{2}[4]{*}{delay} & \multicolumn{2}{c}{Spatial} & \multicolumn{2}{c}{Object} & \multicolumn{2}{c}{Goal} & \multicolumn{2}{c|}{Long} & \multicolumn{2}{c}{Average} \\
    \cmidrule{4-13} & & & SR $\uparrow$ & Steps $\downarrow$ & SR $\uparrow$ & Steps $\downarrow$ & SR $\uparrow$ & Steps $\downarrow$ & SR $\uparrow$ & Steps $\downarrow$ & SR $\uparrow$ & Steps $\downarrow$ \\
    \midrule
    \multirow{17}[10]{*}{Infer} & Sync. & 0     & 96.6  & 107.6 & 98.0  & 137.8 & 95.0  & 117.5 & 96.8  & 260.6 & 96.6  & 155.8 \\
    \cmidrule{2-13}          
    & \multirow{4}[2]{*}{Naive Async.} 
    & 5 & 90.8  & 113.0 & 90.4  & 147.9 & 91.0  & 125.6 & 85.8  & 289.4 & 89.5  & 169.0 \\
    & & 10    & 83.2  & 123.4 & 85.0  & 155.8 & 89.8  & 127.3 & 72.6  & 325.5 & 82.7  & 183.0 \\
    & & 15    & 79.6  & 128.6 & 77.4  & 166.7 & 87.8  & 131.5 & 60.2  & 359.5 & 76.3  & 196.6 \\
    & & 20    & 67.6  & 142.0 & 77.0  & 167.5 & 83.6  & 137.0 & 45.0  & 397.2 & 68.3  & 210.9 \\
    \cmidrule{2-13} 
    & \multirow{4}[2]{*}{TE} 
    & 5 & 91.0  & 114.4 & 88.6  & 150.1 & 93.0  & 122.0 & 78.6  & 305.5 & 87.8  & 173.0 \\
    & & 10    & 87.4  & 118.7 & 84.8  & 155.7 & 92.0  & 124.1 & 73.0  & 321.9 & 84.3  & 180.1 \\
    & & 15    & 81.0  & 126.3 & 81.2  & 161.4 & 90.8  & 125.2 & 69.2  & 330.3 & 80.6  & 185.8 \\
    & & 20    & 74.0  & 134.1 & 77.6  & 167.0 & 88.6  & 128.3 & 59.0  & 358.5 & 74.8  & 197.0 \\
    \cmidrule{2-13} 
    & \multirow{4}[2]{*}{BID} 
    & 5 & 91.6  & 113.0 & 90.6  & 148.3 & 91.4  & 124.6 & 87.4  & 286.9 & 90.3  & 168.2 \\
    & & 10    & 85.8  & 121.4 & 83.4  & 157.9 & 89.8  & 127.2 & 77.0  & 315.7 & 84.0  & 180.5 \\
    & & 15    & 82.0  & 125.0 & 78.4  & 166.4 & 85.4  & 135.8 & 58.6  & 361.9 & 76.1  & 197.3 \\
    & & 20    & 71.4  & 137.7 & 78.4  & 166.5 & 85.2  & 135.3 & 50.2  & 383.8 & 71.3  & 205.8 \\
    \cmidrule{2-13} & \multirow{4}[2]{*}{RTC} 
    & 5 & 91.4  & 113.1 & 89.4  & 149.2 & 91.6  & 124.8 & 87.0  & 286.5 & 89.9  & 168.4 \\
    & & 10    & 84.2  & 122.0 & 84.2  & 156.9 & 90.4  & 127.2 & 73.8  & 320.5 & 83.2  & 181.6 \\
    & & 15    & 80.0  & 128.2 & 78.6  & 166.0 & 88.8  & 130.0 & 62.4  & 352.8 & 77.5  & 194.2 \\
    & & 20    & 71.6  & 137.3 & 79.0  & 165.0 & 86.0  & 134.9 & 58.0  & 364.9 & 73.7  & 200.5 \\
    \midrule
    \multirow{16}[8]{*}{Train.} 
    & \multirow{4}[2]{*}{T-RTC} 
    & 5 & 87.4  & 118.4 & 88.6  & 150.2 & 93.6  & 121.2 & 87.8  & 284.3 & 89.4  & 168.5 \\
    & & 10    & 82.2  & 123.4 & 83.4  & 157.6 & 91.0  & 125.0 & 79.4  & 304.8 & 84.0  & 177.7 \\
    & & 15    & 81.8  & 123.7 & 78.8  & 164.9 & 89.8  & 127.8 & 61.2  & 356.2 & 77.9  & 193.1 \\
    & & 20    & 74.0  & 134.9 & 76.4  & 170.2 & 86.2  & 133.9 & 59.2  & 361.5 & 74.0  & 200.1 \\
    \cmidrule{2-13} 
    & \multirow{4}[2]{*}{VLASH} 
    & 5 & 87.9 & 119.1 & 87.4 & 151.8 & 93.2 & 121.8 & 83.4 & 295.0 & 88.0 & 171.9  \\
    & & 10 & 82.4 & 123.7 & 84.2 & 156.5 & 88.4 & 128.8 & 78.6 & 306.8 & 83.4 & 178.9 \\
    & & 15 & 77.3 & 129.7 & 78.4 & 165.4 & 88.2 & 130.1 & 74.2 & 324.5 & 79.5 & 187.4 \\
    & & 20 & 76.2 & 132.0 & 75.4 & 171.5 & 79.0 & 144.4 & 67.3 & 341.7 & 74.5 & 197.4  \\
    \cmidrule{2-13} 
    & \multirow{4}[2]{*}{REMAC} 
    & 5 & 90.6  & 114.2 & 91.0  & 147.9 & 94.0  & 119.5 & 87.6  & 284.6 & 90.8  & 166.6 \\
    & & 10    & 87.8  & 117.7 & 85.2  & 155.4 & 91.6  & 123.8 & 77.0  & 313.3 & 85.4  & 177.6 \\
    & & 15    & 82.0  & 125.3 & 80.6  & 162.2 & 90.0  & 127.3 & 61.4  & 355.7 & 78.5  & 192.6 \\
    & & 20    & 74.2  & 134.1 & 79.4  & 163.4 & 87.2  & 130.7 & 54.8  & 370.1 & 73.9  & 199.6 \\
    \cmidrule{2-13} 
    & \multirow{4}[2]{*}{FutureRTC} 
    & 5 & 95.0  & 109.6 & 96.2  & 140.4 & 93.6  & 120.5 & 91.8  & 279.1 & 94.2  & 162.4 \\
    & & 10    & 95.0  & 110.8 & 93.8  & 143.3 & 93.4  & 121.5 & 87.4  & 291.9 & 92.4  & 166.9 \\
    & & 15    & 93.2  & 112.5 & 94.0  & 142.7 & 92.2  & 123.0 & 84.4  & 302.0 & 91.0  & 170.1 \\
    & & 20    & 90.4  & 115.8 & 92.6  & 145.8 & 90.8  & 125.3 & 80.2  & 313.7 & 88.5  & 175.1 \\
    \bottomrule
    \end{tabular}}
  \label{tab: libero_pi05}
\end{table*}

\begin{table*}[!t]
  \centering
  \caption{Detailed performance comparison of SmolVLA-450M~\cite{Smolvla} in the LIBERO~\cite{Libero} benchmark with different methods across various inference delays $d\in \{5,10,15,20\}$. }
  \resizebox{\textwidth}{!}{
    \begin{tabular}{c|c|c|cccccccc|cc}
    \toprule
    \multirow{2}[4]{*}{} & \multirow{2}[4]{*}{Methods} & \multirow{2}[4]{*}{delay} & \multicolumn{2}{c}{Spatial} & \multicolumn{2}{c}{Object} & \multicolumn{2}{c}{Goal} & \multicolumn{2}{c|}{Long} & \multicolumn{2}{c}{Average} \\
    \cmidrule{4-13} & & & SR $\uparrow$ & Steps $\downarrow$ & SR $\uparrow$ & Steps $\downarrow$ & SR $\uparrow$ & Steps $\downarrow$ & SR $\uparrow$ & Steps $\downarrow$ & SR $\uparrow$ & Steps $\downarrow$ \\
    \midrule
    \multirow{17}[2]{*}{Infer} 
    & Sync. & 0     & 77.8  & 128.2 & 85.4  & 156.2 & 83.4  & 139.9 & 63.8  & 355.2 & 77.6  & 194.9 \\
    \cmidrule{2-13}
    & \multirow{4}[0]{*}{Naive Async.} 
    & 5 & 71.6  & 135.6 & 75.0  & 168.6 & 81.0  & 142.9 & 53.6  & 378.4 & 70.3  & 206.4 \\
    & & 10    & 68.0  & 139.0 & 71.6  & 173.8 & 77.0  & 150.1 & 41.4  & 409.8 & 64.5  & 218.2 \\
    & & 15    & 65.2  & 142.5 & 71.6  & 173.7 & 72.2  & 157.1 & 36.2  & 423.4 & 61.3  & 224.2 \\
    & & 20    & 55.8  & 154.1 & 72.4  & 172.3 & 67.4  & 164.9 & 29.2  & 440.0 & 56.2  & 232.8 \\
    \cmidrule{2-13}
    & \multirow{4}[0]{*}{TE} 
    & 5 & 69.0  & 138.1 & 76.6  & 165.9 & 79.0  & 145.0 & 49.0  & 386.2 & 68.4  & 208.8 \\
    & & 10    & 66.6  & 140.5 & 76.4  & 166.4 & 78.2  & 146.6 & 45.6  & 398.1 & 66.7  & 212.9 \\
    & & 15    & 59.8  & 148.7 & 75.4  & 167.6 & 75.6  & 151.8 & 39.6  & 411.6 & 62.6  & 220.0 \\
    & & 20    & 54.8  & 154.8 & 72.0  & 173.1 & 74.2  & 153.1 & 37.8  & 415.8 & 59.7  & 224.2 \\
    \cmidrule{2-13}
    & \multirow{4}[0]{*}{BID} 
    & 5 & 73.0  & 134.0 & 75.4  & 167.8 & 80.2  & 144.5 & 51.4  & 384.2 & 70.0  & 207.6 \\
    & & 10    & 67.8  & 139.5 & 71.2  & 174.5 & 78.0  & 147.2 & 42.8  & 404.6 & 65.0  & 216.4 \\
    & & 15    & 65.8  & 141.9 & 71.2  & 174.4 & 73.2  & 156.0 & 36.4  & 419.8 & 61.7  & 223.0 \\
    & & 20    & 53.8  & 156.6 & 72.4  & 172.6 & 68.4  & 162.1 & 33.6  & 426.4 & 57.1  & 229.4 \\
    \cmidrule{2-13}
    & \multirow{4}[1]{*}{RTC} 
    & 5 & 73.4  & 133.8 & 75.0  & 168.6 & 79.6  & 145.0 & 50.6  & 385.0 & 69.7  & 208.1 \\
    & & 10    & 67.8  & 139.5 & 71.2  & 174.5 & 77.2  & 148.7 & 43.2  & 405.3 & 64.9  & 217.0 \\
    & & 15    & 65.6  & 142.1 & 71.0  & 174.6 & 73.8  & 155.3 & 40.0  & 411.7 & 62.6  & 220.9 \\
    & & 20    & 57.6  & 151.1 & 71.2  & 174.3 & 69.2  & 161.5 & 36.6  & 418.2 & 58.7  & 226.3 \\
    \midrule
    \multirow{16}[2]{*}{Train.} 
    & \multirow{4}[1]{*}{T-RTC} 
    & 5 & 74.4  & 133.1 & 71.2  & 175.4 & 77.2  & 151.3 & 56.6  & 375.1 & 69.9  & 208.7 \\
    & & 10    & 72.4  & 134.2 & 67.8  & 180.2 & 76.0  & 153.0 & 46.2  & 395.8 & 65.6  & 215.8 \\
    & & 15    & 67.4  & 140.0 & 66.0  & 182.8 & 74.0  & 155.3 & 39.2  & 410.0 & 61.7  & 222.0 \\
    & & 20    & 63.8  & 144.5 & 66.6  & 181.7 & 69.8  & 161.9 & 37.2  & 418.2 & 59.4  & 226.6 \\
    \cmidrule{2-13}
    & \multirow{4}[0]{*}{VLASH} 
    & 5 & 70.8  & 136.1 & 77.4  & 167.3 & 77.8  & 149.2 & 51.2  & 381.5 & 69.3  & 208.5 \\
    & & 10    & 66.2  & 141.8 & 71.4  & 174.7 & 78.0  & 149.3 & 47.4  & 404.7 & 65.8  & 217.6 \\
    & & 15    & 67.0  & 140.4 & 64.8  & 183.0 & 80.6  & 144.0 & 46.2  & 394.8 & 64.7  & 215.6 \\
    & & 20    & 64.8  & 143.9 & 60.4  & 190.1 & 76.8  & 150.6 & 45.6  & 399.1 & 61.9  & 220.9 \\
    \cmidrule{2-13}
    & \multirow{4}[0]{*}{REMAC} 
    & 5 & 73.0  & 133.8 & 77.4  & 165.5 & 81.2  & 142.2 & 52.0  & 383.4 & 70.9  & 206.2 \\
    & & 10    & 69.4  & 138.0 & 72.2  & 173.1 & 78.0  & 147.5 & 46.4  & 395.5 & 66.5  & 213.5 \\
    & & 15    & 66.0  & 141.6 & 70.8  & 175.2 & 73.2  & 154.6 & 38.8  & 412.1 & 62.2  & 220.9 \\
    & & 20    & 59.6  & 148.6 & 71.6  & 173.7 & 72.4  & 155.9 & 35.2  & 421.4 & 59.7  & 224.9 \\
    \cmidrule{2-13}
    & \multicolumn{1}{c|}{\multirow{4}[1]{*}{FutureRTC}} 
    & 5 & 77.2  & 129.4 & 83.6  & 158.4 & 82.4  & 140.7 & 59.8  & 366.2 & 75.8  & 198.7 \\
    & & 10    & 75.4  & 131.6 & 82.0  & 160.8 & 80.4  & 142.9 & 55.4  & 375.5 & 73.3  & 202.7 \\
    & & 15    & 72.2  & 135.3 & 79.6  & 164.6 & 81.6  & 142.9 & 52.8  & 382.6 & 71.6  & 206.4 \\
    & & 20    & 72.8  & 134.8 & 75.0  & 170.8 & 79.4  & 143.5 & 50.2  & 388.7 & 69.4  & 209.4 \\
    \bottomrule
    \end{tabular}}
  \label{tab: libero_smolvla}%
\end{table*}%

\begin{table*}[!t]
  \centering
    \caption{Detailed ablation results on the LIBERO~\cite{Libero} benchmark across the four suites with various inference delays $d\in\{5,10,15,20\}$, using frozen $\pi_{0.5}$~\cite{pi05} and SmolVLA-450M~\cite{Smolvla}.}
  \resizebox{\textwidth}{!}{
    \begin{tabular}{c|c|cccccccc|cc}
    \toprule
    \multicolumn{12}{c}{\textbf{$\pi_{0.5}$}} \\
    \midrule
    \multirow{2}[4]{*}{Methods} & \multirow{2}[4]{*}{delay} & \multicolumn{2}{c}{Spatial} & \multicolumn{2}{c}{Object} & \multicolumn{2}{c}{Goal} & \multicolumn{2}{c|}{Long} & \multicolumn{2}{c}{Average} \\
    \cmidrule{3-12} & & SR $\uparrow$ & Steps $\downarrow$ & SR $\uparrow$ & Steps $\downarrow$ & SR $\uparrow$ & Steps $\downarrow$ & SR $\uparrow$ & Steps $\downarrow$ & SR $\uparrow$ & Steps $\downarrow$ \\
    \midrule
    \multirow{4}[2]{*}{Baseline} 
    & 5 & 21.8  & 204.1 & 53.0  & 240.4 & 67.2  & 187.2 & 45.6  & 433.5 & 46.9  & 266.3 \\
    & 10 & 2.4   & 218.7 & 1.6   & 279.8 & 36.6  & 240.2 & 6.0   & 512.2 & 11.7  & 312.7 \\
    & 15 & 1.0   & 219.7 & 0.0   & 280.0 & 18.8  & 273.4 & 1.0   & 519.0 & 5.2   & 323.0 \\
    & 20 & 0.2   & 219.9 & 0.2   & 280.0 & 8.2   & 291.5 & 1.2   & 518.7 & 2.5   & 327.5 \\
    \midrule
    \multirow{4}[2]{*}{+SCM} 
    & 5 & 26.0  & 200.8 & 53.8  & 239.4 & 65.4  & 189.6 & 42.8  & 435.5 & 47.0  & 266.3 \\
    & 10 & 2.6   & 219.2 & 2.2   & 279.6 & 38.2  & 237.8 & 5.0   & 513.7 & 12.0  & 312.6 \\
    & 15 & 0.6   & 219.9 & 0.2   & 280.0 & 18.6  & 274.1 & 2.0   & 517.6 & 5.4   & 322.9 \\
    & 20 & 0.4   & 220.0 & 0.4   & 279.8 & 9.2   & 289.7 & 1.6   & 518.4 & 2.9   & 327.0 \\
    \midrule
    \multirow{4}[2]{*}{+OPM} 
    & 5 & 95.0  & 109.6 & 96.0  & 140.5 & 93.8  & 120.0 & 92.4  & 277.0 & 94.3  & 161.8 \\
    & 10 & 93.2  & 111.2 & 92.8  & 143.5 & 92.8  & 121.6 & 84.6  & 300.7 & 90.9  & 169.3 \\
    & 15 & 91.6  & 113.5 & 92.8  & 143.1 & 92.0  & 123.2 & 82.2  & 312.0 & 89.7  & 173.0 \\
    & 20 & 90.2  & 115.5 & 93.8  & 141.9 & 89.4  & 128.5 & 74.6  & 329.3 & 87.0  & 178.8 \\
    \midrule
    \multirow{4}[2]{*}{+$\mathcal{L}_{\text{policy}}$} 
    & 5 & 95.0  & 109.6 & 96.2  & 140.4 & 93.6  & 120.5 & 91.8  & 279.1 & 94.2  & 162.4 \\
    & 10 & 95.0  & 110.8 & 93.8  & 143.3 & 93.4  & 121.5 & 87.4  & 291.9 & 92.4  & 166.9 \\
    & 15 & 93.2  & 112.5 & 94.0  & 142.7 & 92.2  & 123.0 & 84.4  & 302.0 & 91.0  & 170.1 \\
    & 20 & 90.4  & 115.8 & 92.6  & 145.8 & 90.8  & 125.3 & 80.2  & 313.7 & 88.5  & 175.1 \\
    \midrule
    \multicolumn{1}{r}{} & \multicolumn{1}{r}{} &       &       &       &       &       &       &       & \multicolumn{1}{r}{} &       &  \\
    \midrule
    \multicolumn{12}{c}{SmolVLA-450M} \\
    \midrule
    \multirow{2}[4]{*}{Methods} 
    & \multirow{2}[4]{*}{delay} & \multicolumn{2}{c}{Spatial} & \multicolumn{2}{c}{Object} & \multicolumn{2}{c}{Goal} & \multicolumn{2}{c|}{Long} & \multicolumn{2}{c}{Average} \\
    \cmidrule{3-12} & & SR $\uparrow$ & Steps $\downarrow$ & SR $\uparrow$ & Steps $\downarrow$ & SR $\uparrow$ & Steps $\downarrow$ & SR $\uparrow$ & Steps $\downarrow$ & SR $\uparrow$ & Steps $\downarrow$ \\
    \midrule
    \multirow{4}[2]{*}{Baseline} 
    & 5 & 10.6  & 212.0 & 23.6  & 262.0 & 50.6  & 208.0 & 19.6  & 482.7 & 26.1  & 291.2 \\
    & 10 & 3.6   & 219.0 & 0.4   & 280.0 & 24.8  & 257.7 & 4.8   & 514.5 & 8.4   & 317.8 \\
    & 15 & 0.2   & 219.9 & 0.0   & 280.0 & 8.6   & 286.5 & 1.0   & 519.6 & 2.5   & 326.5 \\
    & 20 & 0.0   & 220.0 & 0.0   & 280.0 & 1.6   & 298.5 & 1.0   & 519.4 & 0.7   & 329.5 \\
    \midrule
    \multirow{4}[2]{*}{+SCM} 
    & 5 & 34.2  & 188.5 & 56.2  & 215.2 & 67.8  & 179.9 & 25.4  & 464.8 & 45.9  & 262.1 \\
    & 10 & 11.4  & 213.6 & 16.0  & 273.0 & 34.6  & 240.2 & 10.4  & 501.4 & 18.1  & 307.1 \\
    & 15 & 7.8   & 216.8 & 7.4   & 278.1 & 19.6  & 269.9 & 6.6   & 510.8 & 10.4  & 318.9 \\
    & 20 & 3.2   & 218.7 & 0.8   & 279.8 & 14.2  & 279.2 & 5.0   & 515.8 & 5.8   & 323.4 \\
    \midrule
    \multirow{4}[2]{*}{+OPM} 
    & 5 & 75.4 & 131.3 & 83.2 & 158.8 & 84.2 & 136.8 & 53.4 & 380.5 & 74.1 & 201.8 \\
    & 10 & 73.4 & 133.6 & 78.8 & 165.2 & 81.8 & 141.7 & 47.6 & 395.6 & 70.4 & 209.0 \\
    & 15 & 72.0 & 135.6 & 79.6 & 164.9 & 83.0 & 140.0 & 45.4 & 402.3 & 70.0 & 210.7 \\
    & 20 & 71.8 & 136.2 & 75.0 & 171.5 & 79.6 & 145.6 & 43.4 & 434.4 & 67.5 & 221.9 \\
    \midrule
    \multirow{4}[2]{*}{+$\mathcal{L}_{\text{policy}}$} 
    & 5 & 77.2  & 129.4 & 83.6  & 158.4 & 82.4  & 140.7 & 59.8  & 366.2 & 75.8  & 198.7 \\
    & 10 & 75.4  & 131.6 & 82.0  & 160.8 & 80.4  & 142.9 & 55.4  & 375.5 & 73.3  & 202.7 \\
    & 15 & 72.2  & 135.3 & 79.6  & 164.6 & 81.6  & 142.9 & 52.8  & 382.6 & 71.6  & 206.4 \\
    & 20 & 72.8  & 134.8 & 75.0  & 170.8 & 79.4  & 143.5 & 50.2  & 388.7 & 69.4  & 209.4 \\
    \bottomrule
    \end{tabular}}
  \label{tab: ablation_supp}
\end{table*}

\section{Limitations}\label{sec: limitation}
While FutureRTC substantially improves the robustness of asynchronous VLA execution, it still has several limitations. First, the observation prediction module models visual changes through motion-driven feature transport and residual synthesis. While highly effective in manipulation scenes where the robot is the primary source of environmental change, it may struggle in highly dynamic scenes dominated by independent external agents. Second, our current framework is instantiated on flow-matching VLA policies with latent-space observation prediction. Extending FutureRTC to other policy architectures, such as autoregressive or discrete-diffusion policies, and more extreme inference delays remains an interesting direction for future work.


\end{document}